\documentclass{amsart}
\usepackage{amsaddr}

\usepackage{booktabs}       
\RequirePackage{amsthm,amsmath,amsfonts,amssymb}
\RequirePackage[authoryear]{natbib}
\RequirePackage[colorlinks,citecolor=blue,urlcolor=blue,backref=page,backref=page]{hyperref}
\RequirePackage{graphicx}
\usepackage{subcaption}

\begin{document}

\title[High-dimensional Variational Inference]{Stable Training of Normalizing Flows for High-dimensional Variational Inference}

\author{Daniel Andrade}
\address{School of Informatics and Data Science, Hiroshima University}
\email{andrade@hiroshima-u.ac.jp}

\begin{abstract}
Variational inference with normalizing flows (NFs) is an increasingly popular alternative to MCMC methods. In particular, NFs based on coupling layers (Real NVPs) are frequently used due to their good empirical performance. In theory, increasing the depth of normalizing flows should lead to more accurate posterior approximations. However, in practice, training deep normalizing flows for approximating high-dimensional posterior distributions is often infeasible due to the high variance of the stochastic gradients. 
In this work, we show that previous methods for stabilizing the variance of stochastic gradient descent can be insufficient to achieve stable training of Real NVPs. As the source of the problem, we identify that, during training, samples often exhibit unusual high values. 
As a remedy, we propose a combination of two methods: (1) soft-thresholding of the scale in Real NVPs, and (2) a bijective soft log transformation of the samples. We evaluate these and other previously proposed modification on several challenging target distributions, including a high-dimensional horseshoe logistic regression model. Our experiments show that with our modifications, stable training of Real NVPs for posteriors with several thousand dimensions is possible, allowing for more accurate marginal likelihood estimation via importance sampling. Moreover, we evaluate several common training techniques and architecture choices and provide practical advise for training NFs for high-dimensional variational inference.
\end{abstract}

\maketitle

\section{Introduction}
\label{sec:intro}

During the past decade, variational inference (VI) has been gaining increased attention also in the statistics community \citep{Blei2017}. 
Thanks to its computational efficiency, VI can provide useful approximations to high-dimensional posterior distributions that are computationally too expensive for MCMC methods.
However, simple variational distributions, like the Gaussian distribution, can lead to arbitrarily bad approximations \citep{Zhang2018}. 

As a promising method for increasing the expressibility of the VI approximation, normalizing flows (NFs) have been proposed  \citep{papamakarios2021normalizing}. In particular, normalizing flows based on coupling layers (Real NVP) are known to have good performance for variational inference \citep{dhaka2021challenges,Vaitl2022}. Real NVPs also have the advantage that both sampling and density estimation is computationally efficient, making them the optimal choice for importance sampling. Unfortunately, for high dimensional posterior distributions, VI with NFs is known to suffer from unstable training, 
and therefore most previous works limit their application to posteriors with less than hundred dimensions \citep{dhaka2021challenges, Vaitl2022,Liang2022a}.

As one source of the unstable training, we identify that Real NVPs with increasing depth are prone to produce exponentially large sample values.
This poses a dilemma, since an increase in depth is also known to increase the expressibility of the resulting posterior approximation \citep{Koehler2021}.
As a remedy, we propose to (1) soft-threshold the scaling in Real NVPs, and (2) to pass all samples through a bijective soft log transformation (named LOFT).\footnote{Preliminary results of our work were presented in \citep{Andrade2023}.}
We demonstrate that with these modifications, the NFs' samples that are used for gradient estimation can exhibit considerably lower variance, which consequently ensures good convergence even for high-dimensional target distributions with fat tails. 
In particular, for logistic regression with the horseshoe prior and more than 4000 dimensions we achieve significantly sharper estimates of the marginal likelihood than previous NFs and sequential Monte Carlo (SMC) \citep{Dai2022}.
Orthogonal to this work, we note that normalizing flows and SMC can be effectively combined to further improve the marginal likelihood estimate \citep{arbel2021annealed,Matthews2022}.

Furthermore, in this article, we also scrutinize and evaluate several other subtle modifications in the architecture and training of Real NVPs, like path gradients \citep{Roeder2017} and annealing \citep{Rezende2015}. Our analysis shows that some of these modifications can have a significant impact on approximation accuracy.
We also empirically verify that for training NFs, the evidence lower bound (ELBO) is well suited as training objective due to its high correlation with the accuracy of the marginal likelihood estimate.

\cite{Behrmann2021} prove that Real NVPs are not globally Lipschitz continuous, which can lead to numerical instability of the inverse. As a remedy, \cite{Behrmann2021,Ardizzone2019} propose to restrict the output out of the scalings. 
However, their focus is on the inverse, in particular on the reconstruction of image data. As we show in Section \ref{sec:results}, their proposed restrictions can be sub-optimal for training in variational inference.
On the other hand, the works in \citep{Salmona2022,Hagemann2021} show that multi-modal target distributions can lead to exploding Lipschitz constants, which can be mitigated by changing the base distribution to a multivariate normal mixture model.
However, training with a more complex base distribution is computationally more demanding, and its effectiveness is apparently highly dependent on how close the more complex base distribution matches with the target distribution.
The works in \citep{Jaini2020,Liang2022a} show that a heavy-tailed target distribution cannot be modeled with Real NVPs, unless the base distribution is also changed to a heavy tailed distribution. In general, training with a heavy-tailed base distribution is more challenging due to the increase in variance of the gradient estimates. We show in our experiments, that our proposed modifications enable stable training with a student-t distribution, leading to considerably better estimates of the marginal likelihood.

Another line of work suggests the use of path gradients \citep{Roeder2017,Vaitl2022} which removes the score term from the gradient estimation of the Kullback-Leibler (KL)-divergence, and can lead to considerably lower variance of the gradient estimates. Our experiments confirm the usefulness of path gradients also for high-dimensional posterior approximation. We note that in order to stabilize training, control variates, as proposed in \citep{ranganath2014black}, could in principle also be applied to NFs. However, in practice, due to the high number of parameters of NFs, such control variates are not computationally feasible. 

The remainder of this article is structured as follows.
In Section \ref{sec:background}, we provide backgrounds on variational inference and Real NVPs. 
In Section \ref{sec:explodingSampleValues}, we describe the problem of exponentially large sample values, followed by Section \ref{sec:stableRealNVP}, where we describe our proposed modifications. Moreover, in Section \ref{sec:details_alternative_choices}, we describe several other possible modifications of Real NVPs. 
In Section \ref{sec:experiments_and_models}, we describe all models (target distributions) and methods that are used for marginal likelihood estimation in our experiments. In Section \ref{sec:results}, we report the main results for five different models with 1000 or more dimensions, 
followed by Section \ref{sec:analysis}, where we analyze the impact of different modifications on training and marginal likelihood estimation, and also compare to Hamiltonian Monte Carlo (HMC).
Finally, in Section \ref{sec:conclusion_and_recommendations}, we summarize our results, and provide recommendations for the architecture and training of NFs with VI. The code for reproducing all experiments is available here \url{https://github.com/andrade-stats/normalizing-flows}.

\section{Background on Normalizing Flows and Real NVP} \label{sec:background}

Variational inference with the reverse Kullback-Leibler (KL) divergence minimizes 
\begin{align} \label{eq:reverse_kl}
	\text{KL}(q_{\boldsymbol{\eta}} || p_*) = \mathbb{E}_{q_{\boldsymbol{\eta}}} \Big[ \log \Big( \frac{q_{\boldsymbol{\eta}}(\boldsymbol{\theta})}{p_*(\boldsymbol{\theta})} \Big) \Big] \, ,
\end{align}
where $q_{\boldsymbol{\eta}}$ denotes the variational approximation parameterized by the vector $\boldsymbol{\eta}$, and $p_*$ denotes the target density given by
\begin{align*} 
	p_*(\boldsymbol{\theta}) := \frac{p(\boldsymbol{\theta}, D)}{p(D)} \, , 
\end{align*}
where $p(\boldsymbol{\theta}, D)$ is the density of the joint distribution of model parameters $\boldsymbol{\theta} \in \mathbb{R}^d$ and data $D$, and $p(D)$ is the marginal likelihood.
We therefore have
\begin{align*}
	\text{KL}(q_{\boldsymbol{\eta}} || p_*) 
	&= \mathbb{E}_{q_{\boldsymbol{\eta}}} \Big[ \log \Big( \frac{q_{\boldsymbol{\eta}}(\boldsymbol{\theta})}{p(\boldsymbol{\theta}, D)} \Big) \Big] +  \log p(D) \, .
\end{align*}
%
%
Normalizing flows approximate the posterior as follows:
\begin{align*}
	\mathbf{z} &\sim q_0 \, , \\
	f_{\boldsymbol{\eta}}(\mathbf{z}) &\sim q_{\boldsymbol{\eta}} \, , 
\end{align*}
i.e. we first sample $\mathbf{z}$ from a base distribution $q_0$, and then use $f_{\boldsymbol{\eta}}(\mathbf{z})$ as a sample from the approximate posterior $q_{\boldsymbol{\eta}}$. 
Assuming that $f : \mathbb{R}^d \rightarrow \mathbb{R}^d$ is a smooth bijective function, the density of the variational approximation is given by  
\begin{align*}
	q_{\boldsymbol{\eta}} (f_{\boldsymbol{\eta}}(\mathbf{z})) = q_0 (\mathbf{z})  | \det (J_f (\mathbf{z}) ) |^{-1}  \, , 
\end{align*}
where $J_f (\mathbf{z}) \in \mathbb{R}^{d \times d}$ is the Jacobian (matrix with all partial derivatives) of $f_{\boldsymbol{\eta}}$ evaluated at point $\mathbf{z}$. 

Using the law of the unconscious statistician, we get
\begin{align*}
	\mathbb{E}_{q_{\boldsymbol{\eta}}} \Big[ \log \Big( \frac{q_{\boldsymbol{\eta}}(\boldsymbol{\theta})}{p(\boldsymbol{\theta}, D)} \Big) \Big]  
	= \mathbb{E}_{q_0} \Big[ \log \Big( \frac{q_{\boldsymbol{\eta}}(f_{\boldsymbol{\eta}}(\mathbf{z}))}{p(f_{\boldsymbol{\eta}}(\mathbf{z}), D)} \Big) \Big] 
\end{align*}
which is called the negative ELBO (evidence lower bound) \citep{Blei2017,papamakarios2021normalizing}.


Finally, for minimizing $\text{KL}(q_{\boldsymbol{\eta}} || p_*)$ with respect to the variational parameters $\boldsymbol{\eta}$, we need to evaluate the gradient 
\begin{align*} 
	\frac{\partial}{\partial {\boldsymbol{\eta}}} \text{KL}(q_{\boldsymbol{\eta}} || p_*) &= \frac{\partial}{\partial {\boldsymbol{\eta}}} \mathbb{E}_{q_{\boldsymbol{\eta}}} \Big[ \log \Big( \frac{q_{\boldsymbol{\eta}}(\boldsymbol{\theta})}{p(\boldsymbol{\theta}, D)} \Big) \Big] \\
	&=  \mathbb{E}_{q_0} \Big[   \frac{\partial}{\partial {\boldsymbol{\eta}}} \log \Big( \frac{q_{\boldsymbol{\eta}}(f_{\boldsymbol{\eta}}(\mathbf{z}))}{p(f_{\boldsymbol{\eta}}(\mathbf{z}), D)} \Big) \Big] \, .
\end{align*}
%
Using samples $\mathbf{z}_k \sim q_0$ for $k = 1, \ldots, b$, we can approximate the gradient using
\begin{align} \label{eq:basicGradientEst}
	\frac{\partial}{\partial {\boldsymbol{\eta}}} \text{KL}(q_{\boldsymbol{\eta}} || p_*) 
	&\approx \frac{1}{b} \sum_{k = 1}^{b} \Big[   \frac{\partial}{\partial {\boldsymbol{\eta}}} \log \Big( \frac{q_{\boldsymbol{\eta}}(f_{\boldsymbol{\eta}}(\mathbf{z}_k))}{p(f_{\boldsymbol{\eta}}(\mathbf{z}_k), D)} \Big) \Big] \, .
\end{align}

Alternatively, as originally proposed in \citep{Roeder2017}, the following gradient estimate can have lower variance
\begin{align} \label{eq:pathGradients}
	\frac{1}{b} \sum_{k = 1}^b \Big[  \frac{\partial}{\partial {\boldsymbol{\eta}}} \log \Big( \frac{q_{\boldsymbol{\eta}}(f_{\boldsymbol{\eta}}(\mathbf{z}_j))}{p(f_{\boldsymbol{\eta}}(\mathbf{z}_j), D)} \Big) -  \frac{\partial}{\partial {\boldsymbol{\eta}}} 
	\log q_{\boldsymbol{\eta}}(\boldsymbol{\theta})\Big|_{\boldsymbol{\theta} = f_{\boldsymbol{\eta}}(\mathbf{z}_j)}  \Big] \, , 
\end{align}
This gradient estimator removes the score term from Equation \eqref{eq:basicGradientEst}, and is also referred to as path gradients \citep{Vaitl2022}. The motivation and derivation can be found in the Appendix.

\subsection{Real NVP}

For training we require that $f_{\boldsymbol{\eta}}(\mathbf{z}), f^{-1}_{\boldsymbol{\eta}}(\boldsymbol{\theta})$, and $  \det (J_f (\mathbf{z}) )$ can be evaluated efficiently.\footnote{Note that when using path gradients we need to evaluate the inverse $f^{-1}_{\boldsymbol{\eta}}(\boldsymbol{\theta})$.}
Real NVP (real-valued non-volume preserving) is an architecture framework for $f$ that has these required properties \citep{Dinh2016}.

Let $S_0$ and $S_1$ be a partition of the index set $\{1,2,3, \ldots, d\}$, whereas $|S_0| = |S_1| = \frac{d}{2}$.\footnote{For ease of representation here, we assume that $d$ is even. The requirement that $S_0$ and $S_1$ are of the same size is not essential. In our implementation, we assign all even indices to $S_0$ and all odd indices to $S_1$.}
Define 
\begin{align}  \label{eq:realNVP_sequence}
	f := f_r \circ f_{r-1} \ldots \circ f_1 \, ,
\end{align}
where the functions $f_i : \mathbb{R}^d \rightarrow \mathbb{R}^d$, for $1,2, \ldots r$ are defined as follows. Let $\mathbf{z}^{(i+1)} := f_i(\mathbf{z}^{(i)})$, and define\footnote{The notation $\mathbf{z}_A$ denotes the set $\{z_j | j \in A \}$.}
\begin{equation} 
	\begin{aligned} 
		\mathbf{z}_A^{(i+1)} &:= \mathbf{z}_A^{(i)} \, ,   \\ 
		\mathbf{z}_B^{(i+1)} &:= \mathbf{z}_B^{(i)} \odot \exp(s_i(\mathbf{z}_A^{(i)})) + t_i(\mathbf{z}_A^{(i)}) \, ,   \label{eq:realNVP}
	\end{aligned}
\end{equation}
where 
\begin{equation}  \label{eq:alternation_for_realNVP}
	\begin{aligned} 
		A &:= S_{(i + 1) \, \text{mod} \, 2} \, , \quad \text{and} \\
		B &:= S_{i \, \text{mod} \, 2} \, ,
	\end{aligned}
\end{equation}
and $s_i : \mathbb{R}^{d/2} \rightarrow \mathbb{R}^{d/2}$ and $t_i : \mathbb{R}^{d/2} \rightarrow \mathbb{R}^{d/2}$ are arbitrary functions that are modeled with a neural network. Note that the exponential $\exp$ is taken component-wise, and $\odot$ denotes the Hadamard product.
The vector $\boldsymbol{\eta}$ contains all neural network weights and bias terms.
Note that $ \mathbf{z}^{(1)}$ is sampled from the base distribution $q_0$, which is often assumed to be a normal distribution.
The Equations \eqref{eq:realNVP} are often referred to as an affine coupling layer \citep{Dinh2016,Lee2021,Ishikawa2023}.

Note that our definition of Real NVP  as given in Equations \eqref{eq:realNVP_sequence}, \eqref{eq:realNVP} and \eqref{eq:alternation_for_realNVP}, is the one often used in practice \citep{Dinh2016,Huang2020,Koehler2021}.
Despite its restrictive form, i.e. no permutation between affine coupling layers, and only alternation of the sets $A$ and $B$, as described in Equation \eqref {eq:alternation_for_realNVP},  \cite{Koehler2021} proved that $f \# q_0$ is a universal approximator to any target distribution on a compact set $U \subseteq \mathbb{R}^d$.\footnote{The notation $f \# q_0$ denotes the push-forward of the probability measure $q_0$ by function $f$. For universal approximation in distribution, $r \geq 3$ and weak regularity conditions on the target distribution are necessary.}

Though, in theory, $r \geq 3$ is sufficient for accurate approximation, in practice, often a much larger number of coupling layers $r$ is required \citep{Koehler2021}. 





\section{Exponentially Large Sample Values} \label{sec:explodingSampleValues}

For any dimension $j$, let us denote by $u$ the maximum scaling value of the Real NVP, i.e.
\begin{align*} 
	u := \max_ { i \in \{1, \ldots r\}}  \big( \exp(s_i(\mathbf{z}_A^{(i)})) \big)_j \, .
\end{align*}
Let us denote by $m$ the maximum value that is sampled from $q_0$ or output by any $t_i$, i.e.
\begin{align*} 
	m := \max_{ i \in \{1, \ldots r\}} \{z_j^{(1)}, \big( t_i(\mathbf{z}_A^{(i)})  \big)_j   \} \, .
\end{align*}
Then we have that 
\begin{align*} 
	z^{(r + 1)}_j \leq m \sum_{ i  = 0}^r u^i  \, .
\end{align*}
In other words, the maximum value that $z^{(r + 1)}_j$ can attain is of order $O(m u^r)$.
That means deep Real NVP, with $r$ being large, are prone to produce samples that are large in magnitude.

In Figures \ref{fig:layer_samples_Funnel} (a) and \ref{fig:layer_samples_HorseshoePriorLogisticRegression} (a), for $r \in \{4, 32, 64\}$, we show the (absolute) maximum value of $z^{(r + 1)}_j$ for each iteration during training.
We observe that sometimes very large values are sampled, leading to an unstable estimate of the ELBO estimate that can cause high variance of the gradient estimates in Equations \eqref{eq:basicGradientEst} and \eqref{eq:pathGradients}.


\section{Stabilizing Real-NVP} \label{sec:stableRealNVP}

The analysis from the previous section suggests two strategies for mitigating large sample values:
\begin{itemize}
	\item Placing an upper bound on the scaling factors $\exp(s_i(\mathbf{z}_A^{(i)}))$.
	\item Performing a log-transformation of the samples.
\end{itemize}

\begin{figure*}
	\centering
	\includegraphics[width=0.9\linewidth]{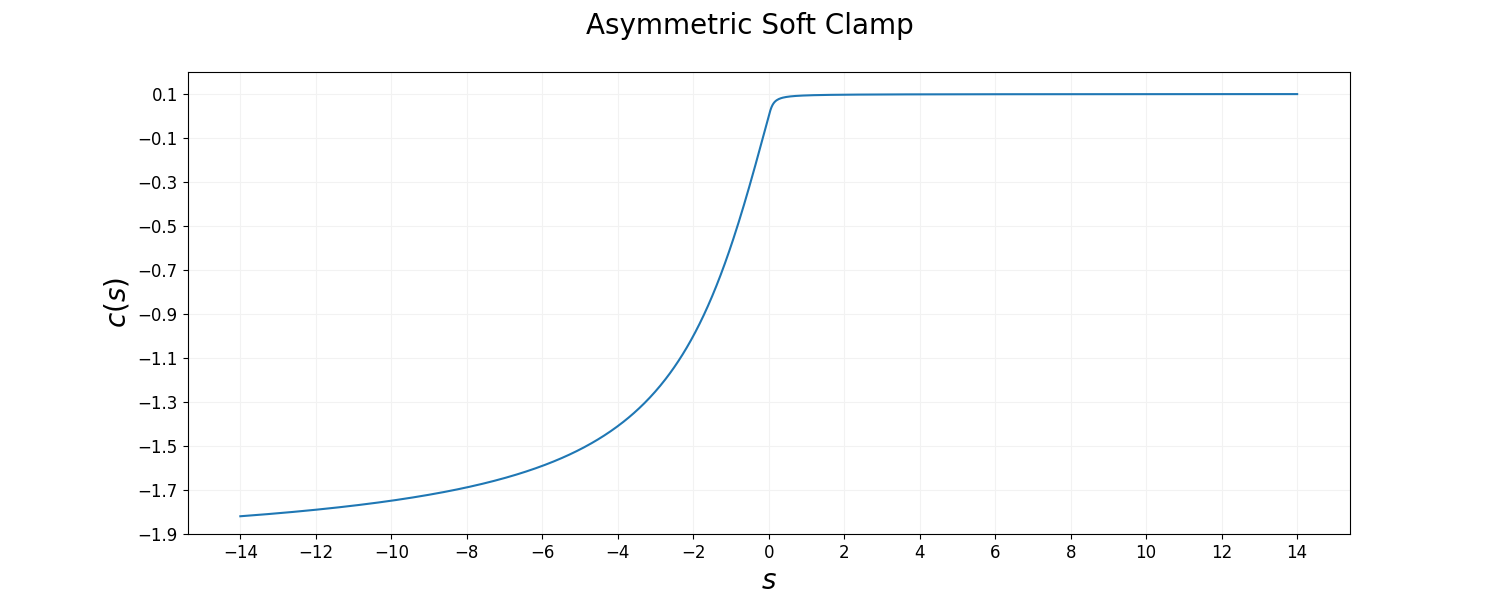}
	\caption{Shows the asymmetric soft clamp function $c(s)$ from Equation \eqref{eq:asym_soft_clamp} with $\alpha_{neg} = 2$ and $\alpha_{pos} = 0.1$.}\label{fig:asym_soft_clamp}
\end{figure*}

\paragraph{Upper and lower bounds on the scaling factors}
We propose to threshold high scaling factors to ensure that $s_i(\mathbf{z}_A^{(i)}) \leq \alpha_{pos}$, where the threshold $\alpha_{pos}$ is set to a fixed small value. 
In order to ensure the stable calculation of the inverse $f^{-1}$ it is also necessary to ensure that $\exp(s_i(\mathbf{z}_A^{(i)}))$ from Equation \eqref{eq:realNVP} does not get too close to zero. We therefore place an additional threshold that enforces $s_i(\mathbf{z}_A^{(i)}) \geq - \alpha_{neg}$. 
Rather than hard thresholding, we found empirically that the following asymmetric soft clamping improved training 
\begin{align} \label{eq:asym_soft_clamp}
	\text{c}(s) &=  \frac{2}{\pi} \begin{cases}
		\alpha_{pos}\arctan(s / \alpha_{pos}) & \text{if } s \geq 0,\\
		\alpha_{neg}\arctan(s / \alpha_{neg}) & \text{if } s < 0 \, .
	\end{cases} 
\end{align}
with $\alpha_{neg} < \alpha_{pos}$. Note that $\text{c}(s)$ is differentiable everywhere.  For our experiments we set $\alpha_{neg} = 2$ and $\alpha_{pos} = 0.1$.\footnote{Overall good performance in terms of ELBO, though manually tuning for each task might improve results further.}
Note that this is a modification of the soft-clamping proposed in \citep{Ardizzone2019}, with the important difference that high-values of $s$ are suppressed more forcefully. The resulting function is shown in Figure \ref{fig:asym_soft_clamp}.
With the proposed asymmetric soft clamping, Equation \eqref{eq:realNVP} changes to 
\begin{align}  \label{eq:realNVP_asym_soft_clamp}
	\mathbf{z}_B^{(i+1)} &:= \mathbf{z}_B^{(i)} \odot \exp(c(s_i(\mathbf{z}_A^{(i)}))) + t_i(\mathbf{z}_A^{(i)}) \, .
\end{align}

\paragraph{Log-transformation of the samples}

\begin{figure*}
	\centering
	\includegraphics[width=0.9\linewidth]{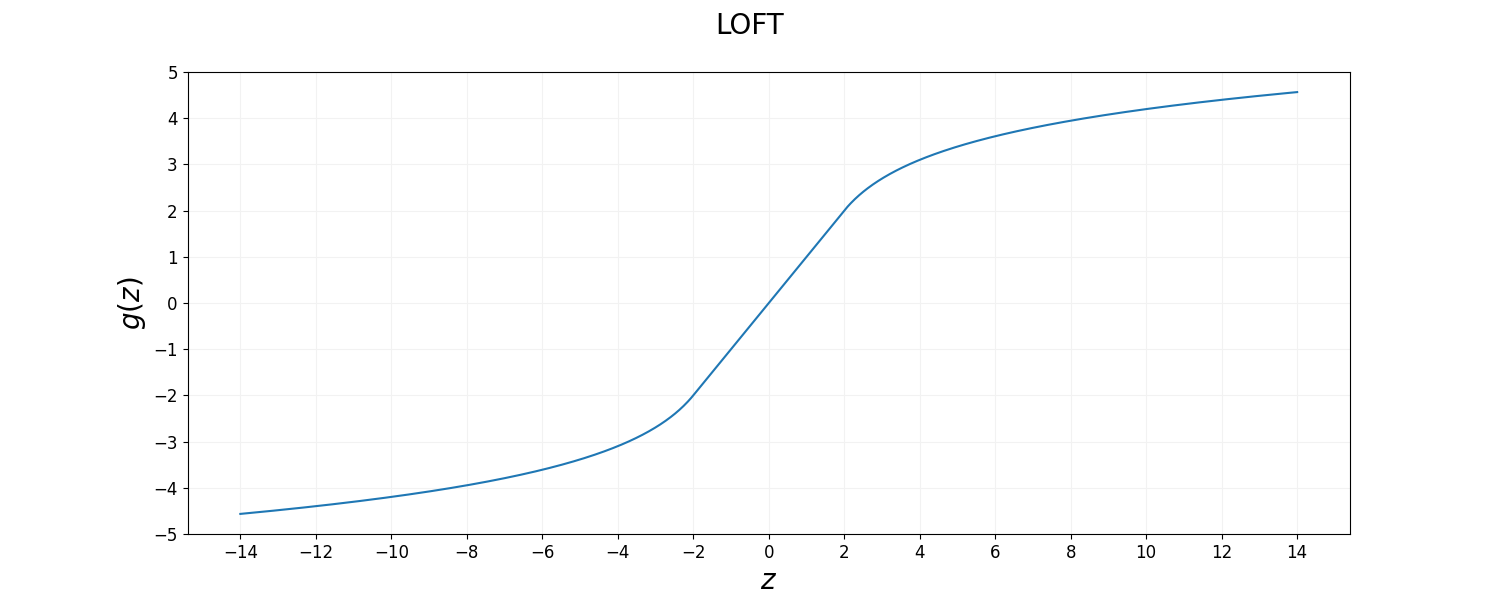}
	\caption{Shows the LOFT function $g(z)$ from Equation \eqref{eq:LOFT} with $\tau = 2$.} \label{fig:LOFT}
\end{figure*}

Since the maximum value of a sample $z^{(r + 1)}$ is of order $O(m u^r)$, it is natural to consider some type of log-transformation $g(z)$ at the final layer. That means instead of \eqref{eq:realNVP_sequence}, we use
\begin{align}  \label{eq:realNVP_sequence_with_LOFT}
	f := g \circ f_r \circ f_{r-1} \ldots \circ f_1 \, .
\end{align}
Note that the function $g$ must be a differentiable bijective function. In particular, 
we propose the following log soft extension (LOFT) layer:
\begin{align} \label{eq:LOFT}
	g(z) &=  \begin{cases}
		\tau + \log(z - \tau + 1) & \text{if } \theta \geq \tau,\\
		-\tau - \log(- z - \tau + 1) & \text{if } \theta \leq -\tau,\\
		z & \text{else}.
	\end{cases} \\
	&= \text{sign}(z) \Big( \log \big( \max( |z| - \tau, 0) + 1 \big)  + \min( |z|, \tau) \Big) \, . \nonumber
\end{align}
The function is shown in Figure \ref{fig:LOFT}: within the range $[-\tau, \tau]$ the layer performs an identity mapping, and outside the range, the absolute value of the function grows only logarithmically; $\tau$ is a fixed pre-specified parameter. 
For a vector $\mathbf{z}$ we apply the  LOFT function element-wise.
%
%
Note that LOFT is a one-to-one function and 
\begin{align*} 
	g^{-1}(z) &=  \text{sign}(z) \Big( \exp \big( \max( |z| - \tau, 0) \big) - 1  + \min( |z|, \tau) \Big)  \, , \\
	\log \big( \frac{\partial}{\partial z} g(z) \Big) &=  - \log \big(  \max( |z| - \tau, 0) + 1 \big) \, .
\end{align*}
Therefore, all necessary calculations can be expressed using only computationally efficient elementary operations (without if-clauses).
For all of our experiments we set $\tau = 100$.

\section{Details on Architecture and Alternative Choices} \label{sec:details_alternative_choices}

\paragraph{Base Distribution and Clipping}
While we propose to use the clipping from Equation \eqref{eq:asym_soft_clamp}, the works in \citep{Ardizzone2019} propose to use symmetric clipping with the $\arctan$ function, whereas \citep{Jaini2020} suggests to use the $\tanh$ function. 
For the base distribution $q_0$ there are two common choices:
\begin{itemize}
	\item  independent identical standard Gaussian distribution for each dimension, i.e. $\prod_{j = 1}^d N(0,1)$. 
	\item independent non-identical student-t distribution $\prod_{j = 1}^d \text{StudentT}(\nu_j)$, where $\text{StudentT}(\nu_j)$ denotes the standard student-t distribution with $\nu_j$ degrees of freedom. 
\end{itemize}
An overview of some variations of Real NVP are given in Table \ref{tab:overview_previous_RealNVP}.

\paragraph{Affine Transformation} 
All previous works transform the base distribution by an  affine transformation
\begin{align*} 
a(\mathbf{z}) = \boldsymbol{\sigma} \odot \mathbf{z} + \boldsymbol{\mu} \, ,
\end{align*}
with trainable parameters $\boldsymbol{\sigma} \in \mathbb{R}_+^d$ and $\boldsymbol{\mu} \in \mathbb{R}^d$.
%
The resulting normalizing flow therefore becomes
\begin{align*}  
f := f_r \circ f_{r-1} \ldots \circ f_1 \circ a \, .
\end{align*}
%
%
When using the modifications from Section \ref{sec:stableRealNVP}, we perform the affine transformation at the end, i.e.
\begin{align*}  
f := a \circ g \circ f_r \circ f_{r-1} \ldots \circ f_1  \, ,
\end{align*}
which has the advantage of re-adjusting the scaling that might have been reduced by LOFT and the asymmetric soft clamping of $s_i$.


\begin{table*}
\centering
\caption{Overview of previously proposed variations of Real NVP.} \label{tab:overview_previous_RealNVP}
\footnotesize
\begin{tabular}{llll}
	\toprule 
	\bfseries Name & \bfseries Base Distribution & \bfseries Clipping & \bfseries Reference\\
	standard & Gaussian & none &  \citep{Dinh2016} \\
	ATAF  & StudentT & tanh &\citep{Jaini2020,Liang2022a} \\
	SymClip & Gaussian & arctan & \citep{Ardizzone2019} \\
	\bottomrule 
\end{tabular}
\end{table*}

\section{Experiments} \label{sec:experiments_and_models}

In this section, we first describe the different probabilistic models that we use for evaluation (Section \ref{sec:targetDistributions_allModels}), and 
then describe the training of the normalizing flows and other MCMC-based methods that are used for comparison (Section \ref{sec:descriptionVI_MCMC}).


\subsection{Models} \label{sec:targetDistributions_allModels}

For evaluation, we use four different target distributions, for which the log marginal likelihood is known (Basic Models),
and a challenging Horseshoe Logistic Regression model with synthetic and real data.

\subsubsection{Basic Models} \label{sec:targetDistributions_basicModels} 

\paragraph{Funnel Distribution}

The funnel distribution, as introduced in \citep{Neal2003}, is given by
\begin{displaymath}
	p_*(\theta_1) =  N(0, 9), \quad  \forall j \in {2,\ldots, d}  : p_*(\theta_j  | \theta_1) =  N(0, e^{\theta_1}) \, .
\end{displaymath}
The target distribution is $p_*(\boldsymbol{\theta})$ with $\boldsymbol{\theta} := (\theta_1, \theta_2, \ldots, \theta_d)  \in \mathbb{R}^d$.

\paragraph{Multivariate Student-T Distribution}
The multivariate student-t distribution with mean $\mathbf{0}$, degrees of freedom $\nu$, and scale matrix $\Sigma$ is given by
\begin{displaymath}
	p_*(\boldsymbol{\theta}) =  \frac{\Gamma((\nu + d) / 2)}{\Gamma(\nu / 2) (\nu \pi)^{d/2} |\Sigma|^{1/2}} \Big( 1 + \frac{1}{\nu}\boldsymbol{\theta}^T \Sigma^{-1} \boldsymbol{\theta}  \Big)^{-(\nu + d) / 2} \, ,
\end{displaymath}
where we set the scale matrix to $\Sigma_{i,j} = 0.8$, for $i \neq j$, and $\Sigma_{i,i} = 1$.

\paragraph{Multivariate Gaussian Mixture}

\begin{displaymath}
	p_*(\boldsymbol{\theta}) =  \sum_{j = 1}^k \frac{1}{k} N(\boldsymbol{\theta}| \boldsymbol{\mu}_j, I_d)  \, ,
\end{displaymath}
where $I_d$ denotes the identity matrix in $\mathbb{R}^{d \times d}$.
Here, we set $k = 3$, and $ \boldsymbol{\mu}_1 = \frac{1}{\sqrt{d}} \mathbf{6}$, $ \boldsymbol{\mu}_2 = - \frac{1}{\sqrt{d}} \mathbf{6}$, and $\boldsymbol{\mu}_3 = \mathbf{0}$ (i.e. in all dimensions constant values $\frac{1}{\sqrt{d}} 6$, $-\frac{1}{\sqrt{d}} 6$, and $0$, respectively).

\paragraph{Conjugate Linear Regression}

We consider the following Bayesian linear regression model
\begin{align*}
	& \sigma^2 \sim \text{Inv-Gamma}(0.5, 0.5) \\
	& \boldsymbol{\beta} \sim N(\mathbf{0}, \sigma^2 I_{d'-1}) \\
	& y_i \stackrel{i.i.d}{\sim} N(\mathbf{x}_i^T \boldsymbol{\beta}, \sigma^2 )  \, \quad \text{for $i \in \{1, \dots n\}$} \, .
\end{align*}
The parameters are $\boldsymbol{\theta} := ( \boldsymbol{\beta}, \sigma^2)$, and the target distribution is the posterior $p(\boldsymbol{\theta} | \mathbf{y}, X)$.
Note that, different from before, $\boldsymbol{\theta} \in \mathbb{R}^{d'} \otimes \mathbb{R}_+$. 
For the normalizing flows, in order to ensure the positiveness for $\sigma^2$, we use the softplus-transformation (as suggested, for example, in \cite{Kucukelbir2017}).

Due to the conjugacy of the priors, the log marginal likelihood has a closed form given by \cite{Chipman2001}: 
\begin{align*}
	\log p(\mathbf{y} | X) = \frac{\Gamma((1 + n) / 2)}{\Gamma(1 / 2) \pi^{n/2} |\Sigma|^{1/2}} \Big( 1 + \mathbf{y}^T \Sigma^{-1}  \mathbf{y} \Big)^{-(1 + n) / 2} \, ,
\end{align*}
where $\Sigma := (I_n - X(X^T X + I_d)^{-1}X^T)^{-1}$, and $X \in \mathbb{R}^{n \times d'}$ contains all explanatory variables.

\subsubsection{Horseshoe Logistic Regression Model} 

We consider the following high-dimensional logistic regression model with the Horseshoe prior \citep{Carvalho2010}: 
\begin{align*} 
	&\tau \sim C^+(0, 1)  \\
	&\text{for $j \in \{1, \dots d'\}$:} \\
	&\quad  \lambda_j \sim C^+(0, 1) \\
	&\quad  \beta_j \sim N(0, \tau^2 \lambda_j^2) \\
	& \mu \sim C^+(0, 10) \\
	& y_i \stackrel{i.i.d}{\sim} \text{Bernoulli}(\sigma(\mathbf{x}_i^T \boldsymbol{\beta} + \mu))  \; \text{, for $i \in \{1, \dots n\}$} \, ,
\end{align*}
where $C^+(0, s)$ is the half-Cauchy distribution with scale $s$, and $\sigma$ is the sigmoid function.
Note that here the total number of parameters $d$ is $2 d' + 2$. Again, to ensure the positiveness for $\tau, \boldsymbol{\lambda}$, and $\mu$, we use the softplus-transformation.

\subsection{Description of Variational Inference and MCMC Methods} \label{sec:descriptionVI_MCMC}

We compare our proposed modifications of Real NVP (\emph{proposed}) to three other variations, namely \emph{standard}, \emph{ATAF} and \emph{SymClip}, as described in Table \ref{tab:overview_previous_RealNVP}. Furthermore, we also compare to mean field Gaussian variational inference.
We did not compare to other normalizing flows like residual flows or continuous flows  \citep{papamakarios2021normalizing}, since, in preliminary experiments, they produced inferior results, or do not allow for importance sampling to estimate the marginal likelihood.

For estimating the marginal likelihood, we also compare to a sequential monte carlo (SMC) sampler. 
SMC is an extension of annealed importance sampling \citep{Neal2001}, but with improved performance for estimating the normalization constant \citep{arbel2021annealed}. Indeed, removing the resampling step in SMC corresponds to annealed importance sampling \citep{Dai2022}.

For all variational inference (VI) methods (normalizing flows and mean field Gaussian), we used Adam  \citep{kingma2015adam} with a mini-batch of size 256 and path gradients (i.e. $b = 256$, from Equation \eqref{eq:pathGradients}) and a fixed learning rate of $10^{-4}$.
The number of training iterations is set to $60,000$, except for the gene expression data experiment where we used $400,000$ iterations.
For SMC, we used a geometric annealing scheme with $100,000$ (intermediate) temperatures.

We also compared to an Hamiltonian Monte Carlo (HMC) sampler \citep{Neal2011} with NUTS \citep{Hoffman2014} in terms of Wasserstein distance to the true target distribution.

\paragraph{Estimation of ELBO and Marginal Likelihood}

Like mean field VI, Real NVPs have the big advantage that sampling and estimation of the density $q_{\boldsymbol{\eta}}$ is fast.
This allows us to use importance sampling (IS) to get an estimate of the marginal likelihood:
\begin{align*} 
	p(D) &= \mathbb{E}_{q_{\boldsymbol{\eta}}} \Big[ \frac{p(\boldsymbol{\theta}, D)}{q_{\boldsymbol{\eta}}(\boldsymbol{\theta})} \Big] \\
	&\approx \frac{1}{b_{eval}} \sum_{k = 1}^{b_{eval}} \frac{p(\boldsymbol{\theta}_k, D)}{q_{\boldsymbol{\eta}}(\boldsymbol{\theta}_k)} \, , 
\end{align*}
%
%
where $\boldsymbol{\theta}_k \sim q_{\boldsymbol{\eta}}$ for $k = 1, \ldots, b_{eval}$. For the final evaluation of ELBO and the IS estimate, we use $20,000$ samples (i.e. $b_{eval} = 20,000$), and repeat each evaluation $20$ times to estimate the Monte Carlo error \citep{koehler2009assessment}.

\section{Results} \label{sec:results}

We evaluated all methods on the models described in \ref{sec:targetDistributions_allModels} for 
different number of dimensions $d$ 
ranging from $d = 10$ to $d = 4002$. For $d < 300$, most methods performed similarly, therefore, we focus here on the results for $d \geq 1000$. The results for $d < 300$ can be found in the Appendix in Tables \ref{tab:ELBO_comparison_basic_models_r64}, \ref{tab:IS_comparison_basic_models_r64}, and \ref{tab:ELBO_and_IS_evaluation_horeshoe_synthetic_r64_r16}.
We show the results for four different variations of our proposed method using either a student-t or Gaussian base distribution, and using either LOFT or not.

\subsection{Basic Models}

\begin{figure}[t]
	\begin{subfigure}{.48\textwidth}
		\centering
		\includegraphics[width=1.0\linewidth,page=1]{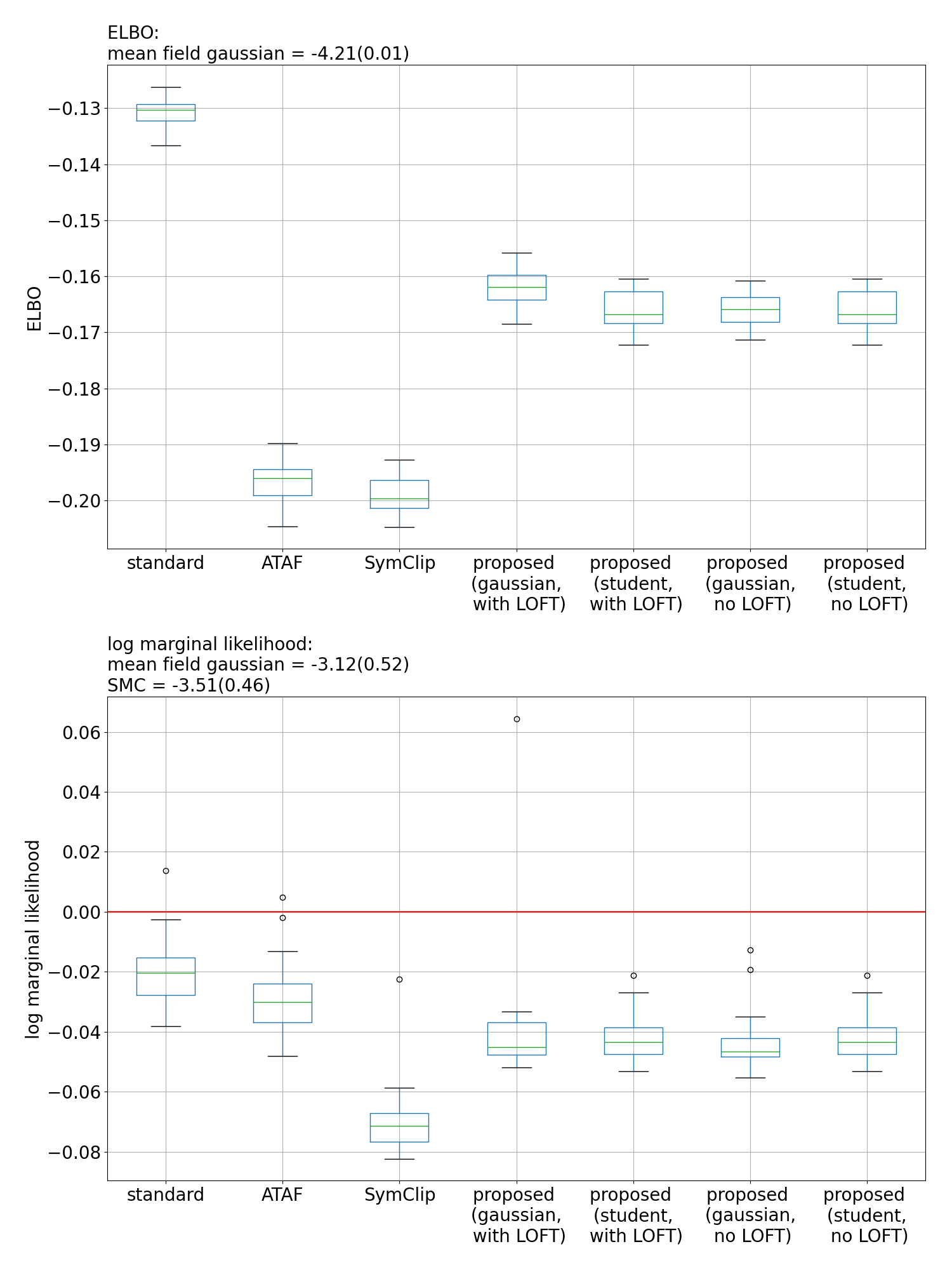} 
		\caption{Funnel}
	\end{subfigure}  \quad
	\begin{subfigure}{0.48\textwidth}
		\centering
		\includegraphics[width=1.0\linewidth,page=1]{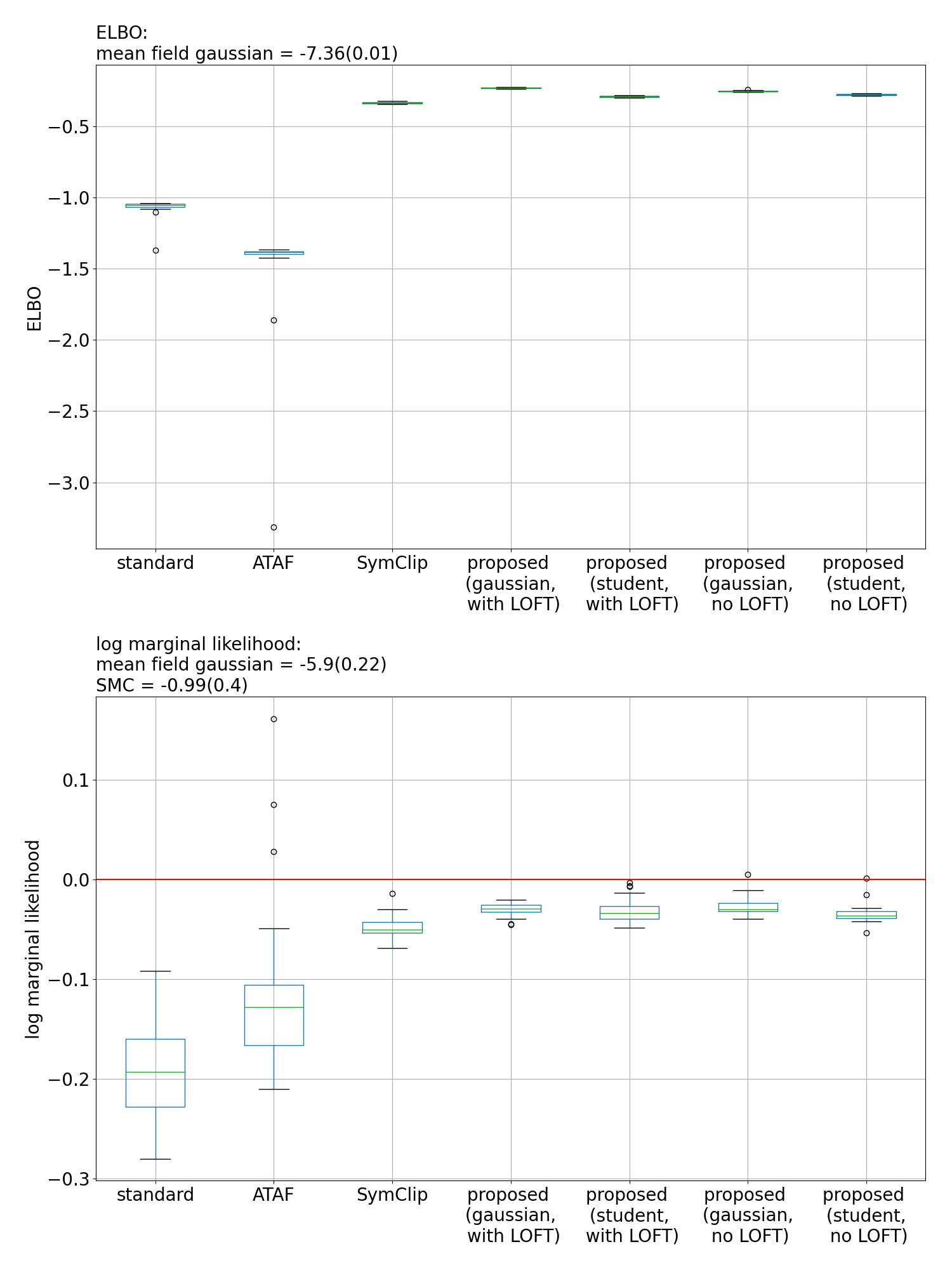}
		\caption{Multivariate Student-T}
	\end{subfigure} \\
	\caption{Shows the ELBO and the log marginal likelihood estimate using importance sampling ($d = 1000$). Red line shows true log marginal likelihood.}
	\label{fig:ELBO_and_MarginalLikelihoodError_all_standard_models_1}
\end{figure}

\begin{figure}[t]
 		\begin{subfigure}{.48\textwidth}
 				\centering
 				\includegraphics[width=1.0\linewidth,page=1]{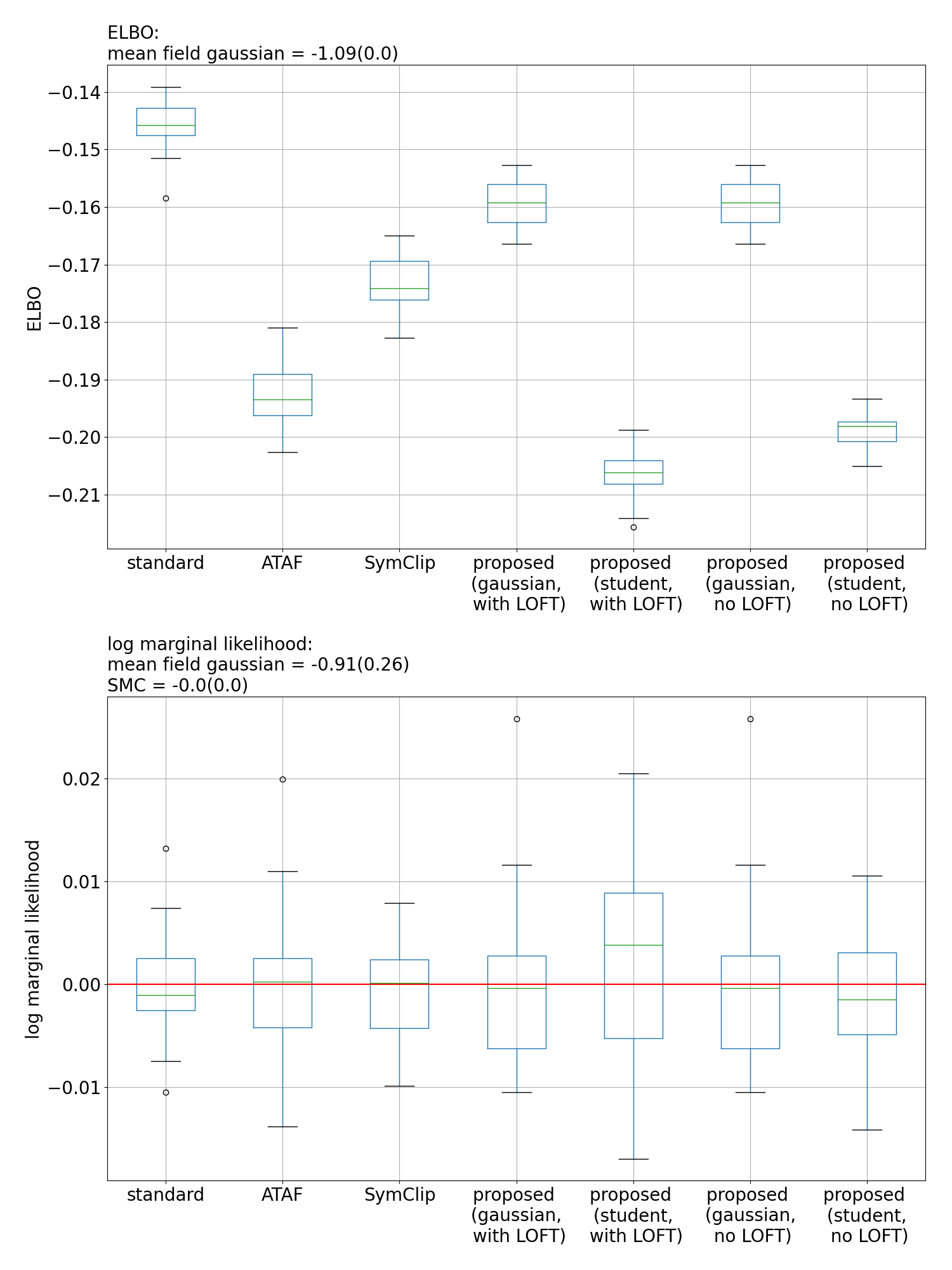}
 				\caption{Multivariate Normal Mixture}
 			\end{subfigure} \quad
 		\begin{subfigure}{.48\textwidth}
 				\centering
 				\includegraphics[width=1.0\linewidth,page=1]{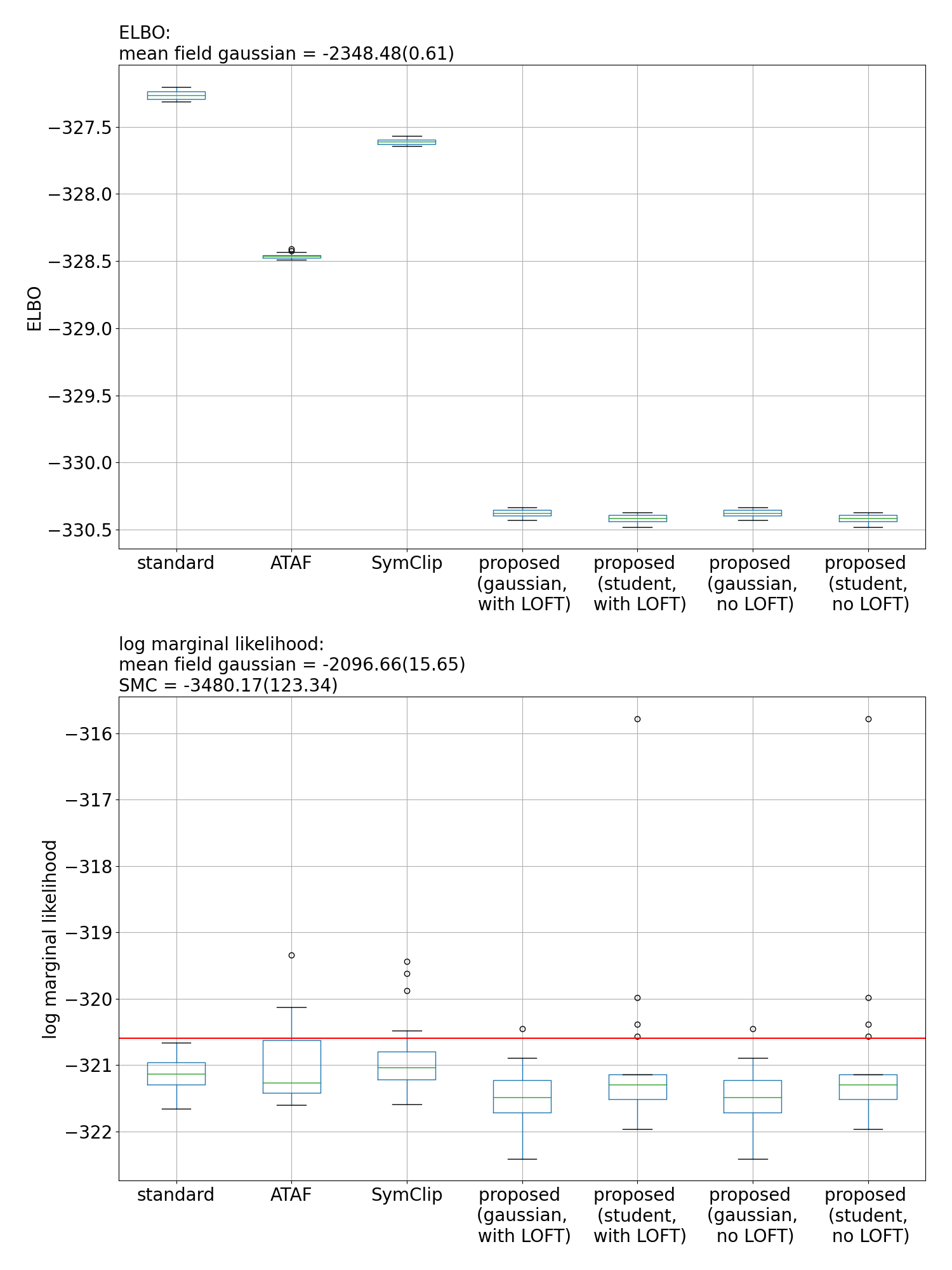}
 				\caption{Conjugate Linear Regression}
 			\end{subfigure}
 	\caption{Shows the ELBO and the log marginal likelihood estimate using importance sampling ($d = 1000$, 
 		for the Conjugate Linear Regression model $d = 1001$). Red line shows true log marginal likelihood.}
 	\label{fig:ELBO_and_MarginalLikelihoodError_all_standard_models_2}
 \end{figure}

Figures \ref{fig:ELBO_and_MarginalLikelihoodError_all_standard_models_1} and \ref{fig:ELBO_and_MarginalLikelihoodError_all_standard_models_2}  show the results for the four basic models, where the true marginal likelihood is known. We see that all methods tend to underestimate the true marginal likelihood.
In most cases, the results of the marginal likelihood estimates based on SMC and the mean field Gaussian approximation (for which average and standard deviation are shown on top of each graph) were considerably worse than the estimates based on normalizing flows. 
For example, for Funnel, the mean field Gaussian approximation and SMC had an error of more than $3$, whereas the error for the normalizing flows was around $0.05$ or less. Overall all normalizing flows performed similarly, except for the challenging multivariate student-t model (Figure \ref{fig:ELBO_and_MarginalLikelihoodError_all_standard_models_1}, (b)), where the proposed method performed slightly better than other normalizing flows. 

\begin{figure}[t]
 	\includegraphics[width=1.0\linewidth,page=1]{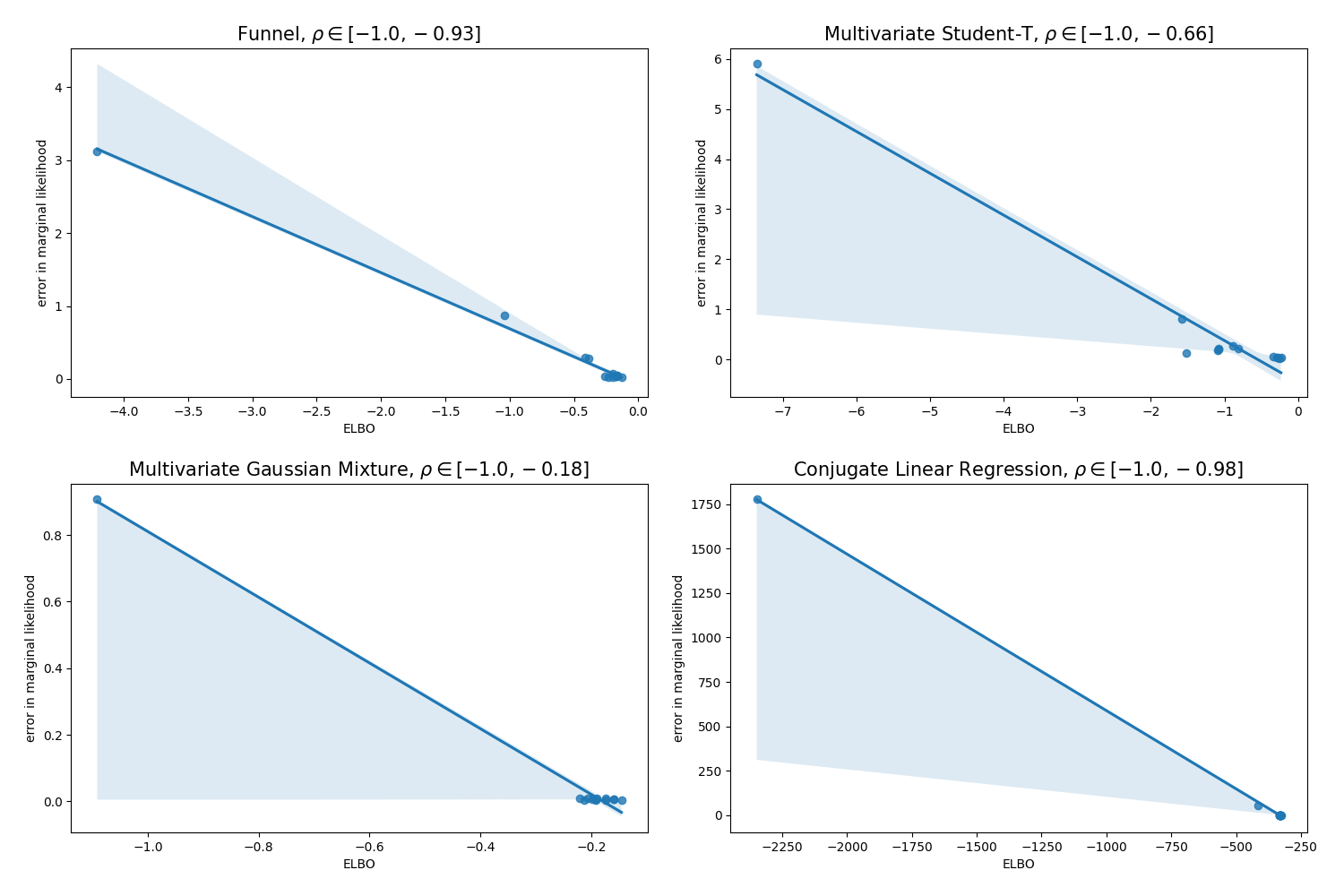}
 	\caption{Plot of ELBO vs error in marginal likelihood of 13 different variational inference methods/models for each model; also shows 95\% confidence level of pearson correlation $\rho$ estimated with bootstrapping.}
 	\label{fig:ELBO_vs_MarginalLikelihoodError}
 \end{figure}

 A higher ELBO tends to improve the estimate of the marginal likelihood.
 In deed, as we show in Figure \ref{fig:ELBO_vs_MarginalLikelihoodError}, optimizing the ELBO is strongly correlated with better estimates of the marginal likelihood that are found via importance sampling.
 
 We note that, in theory, training normalizing flows with the forward KL \citep{jerfel2021variational} should lead to better estimates of the marginal likelihood via importance sampling. However, in practice, we found that the ELBO, of the reverse KL, is much more stable in training and, as suggested by Figure \ref{fig:ELBO_vs_MarginalLikelihoodError}, highly correlated to better estimates of the marginal likelihood.
 
\subsection{Horseshoe Logistic Regression Model} 

\begin{figure*}
	\centering
	\includegraphics[width=0.7\linewidth]{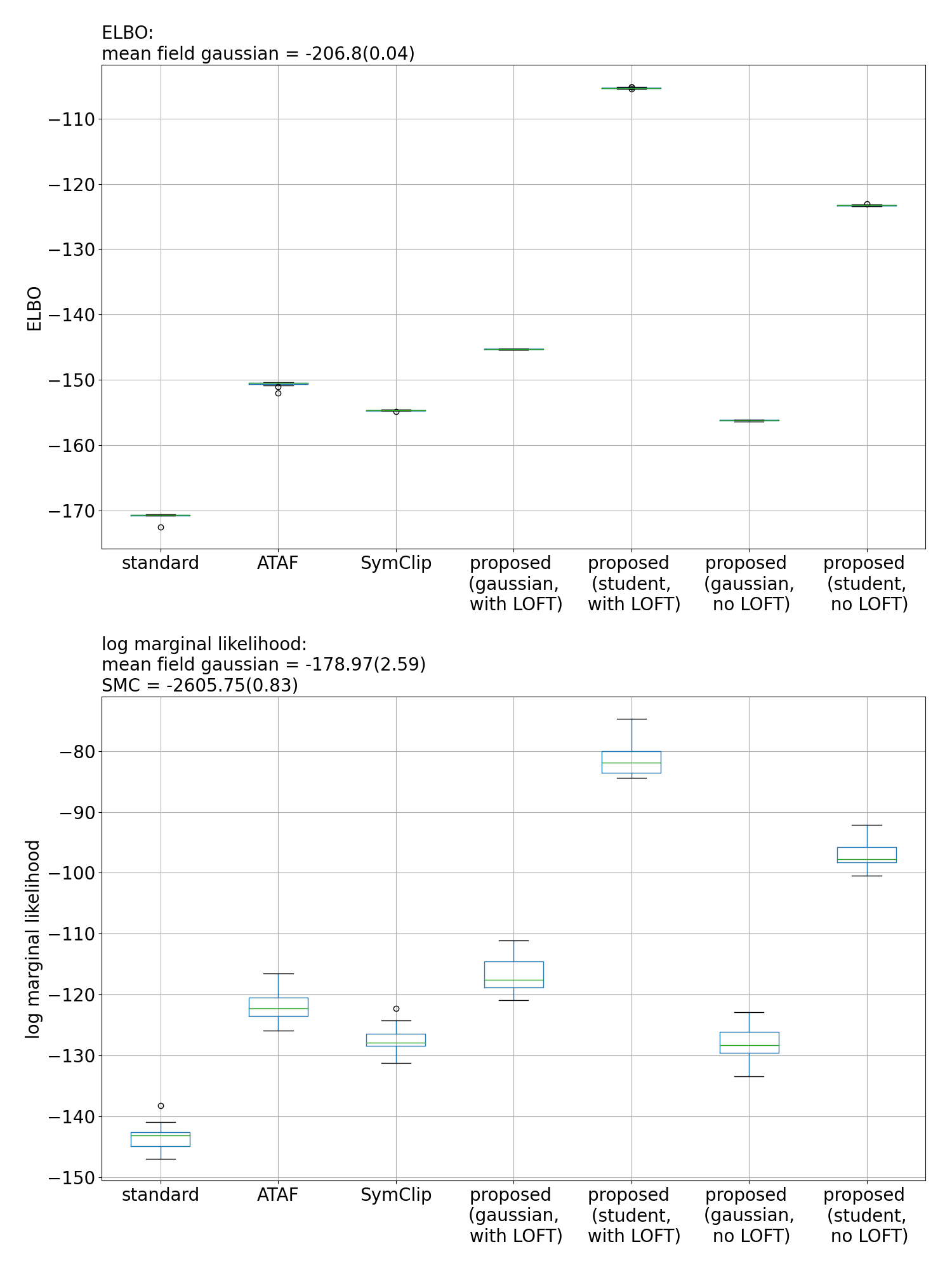}
	\caption{Shows the ELBO and the log marginal likelihood estimate using importance sampling for the Horseshoe logistic regression models on synthetic data with $d = 2002$.}
	\label{fig:ELBO_and_MarginalLikelihoodError_HorseshoePriorLogisticRegression_synthetic}
\end{figure*}

\begin{figure*}
	\centering
	\includegraphics[width=0.7\linewidth]{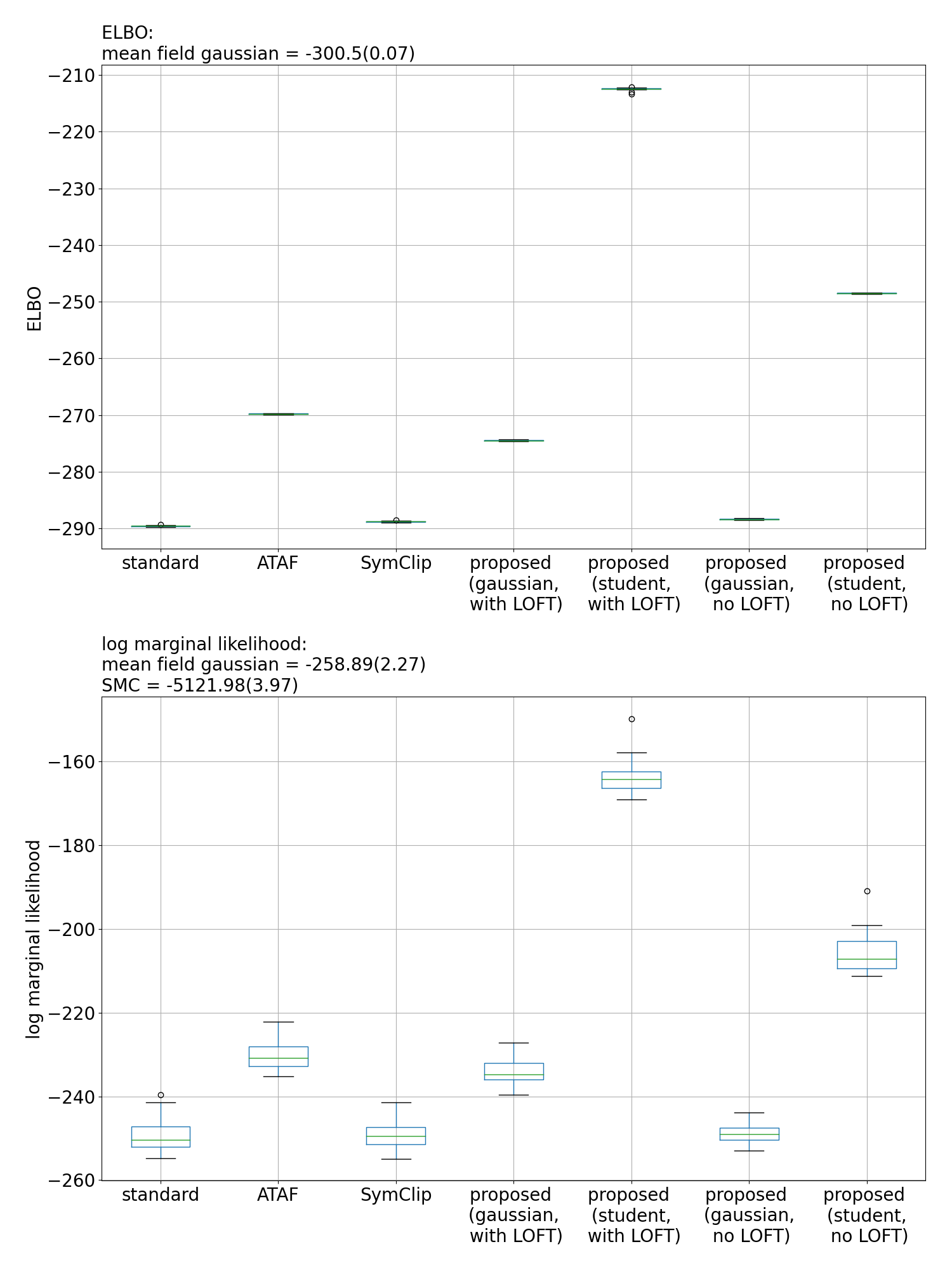}
	\caption{Shows the ELBO and the log marginal likelihood estimate using importance sampling for the Horseshoe logistic regression models on colon data with $d = 4002$.}
	\label{fig:ELBO_and_MarginalLikelihoodError_HorseshoePriorLogisticRegression_colon}
\end{figure*}

Next, in Figure \ref{fig:ELBO_and_MarginalLikelihoodError_HorseshoePriorLogisticRegression_synthetic}, we show the results for the Horseshoe logistic regression model on synthetic data with $d = 2002$.
Finally, we also evaluate all methods on a gene expression data set with 2000 genes from 62 tissue samples either belonging to tumor or normal colon tissue \citep{Alon1999};  
the estimated marginal likelihoods of the Horseshoe Logistic Regression model ($d = 4002$) are shown in Figure \ref{fig:ELBO_and_MarginalLikelihoodError_HorseshoePriorLogisticRegression_colon}. 
As before, mean field Gaussian VI and SMC fail due to the high dimensionality. For lower dimensions, as shown in the Appendix in Table \ref{tab:ELBO_and_IS_evaluation_horeshoe_synthetic_r64_r16}, both methods perform better. Among all normalizing flows the proposed method with a student-t as base distribution and LOFT, performs best. Compared to ATAF, for the synthetics data, the proposed method improves ELBO by around 30\% and for the colon tissue data by around 20\%, resulting in considerably higher estimates of the marginal likelihood.

\section{Analysis} \label{sec:analysis}

\begin{figure*}
	\centering
    \includegraphics[width=0.7\linewidth]{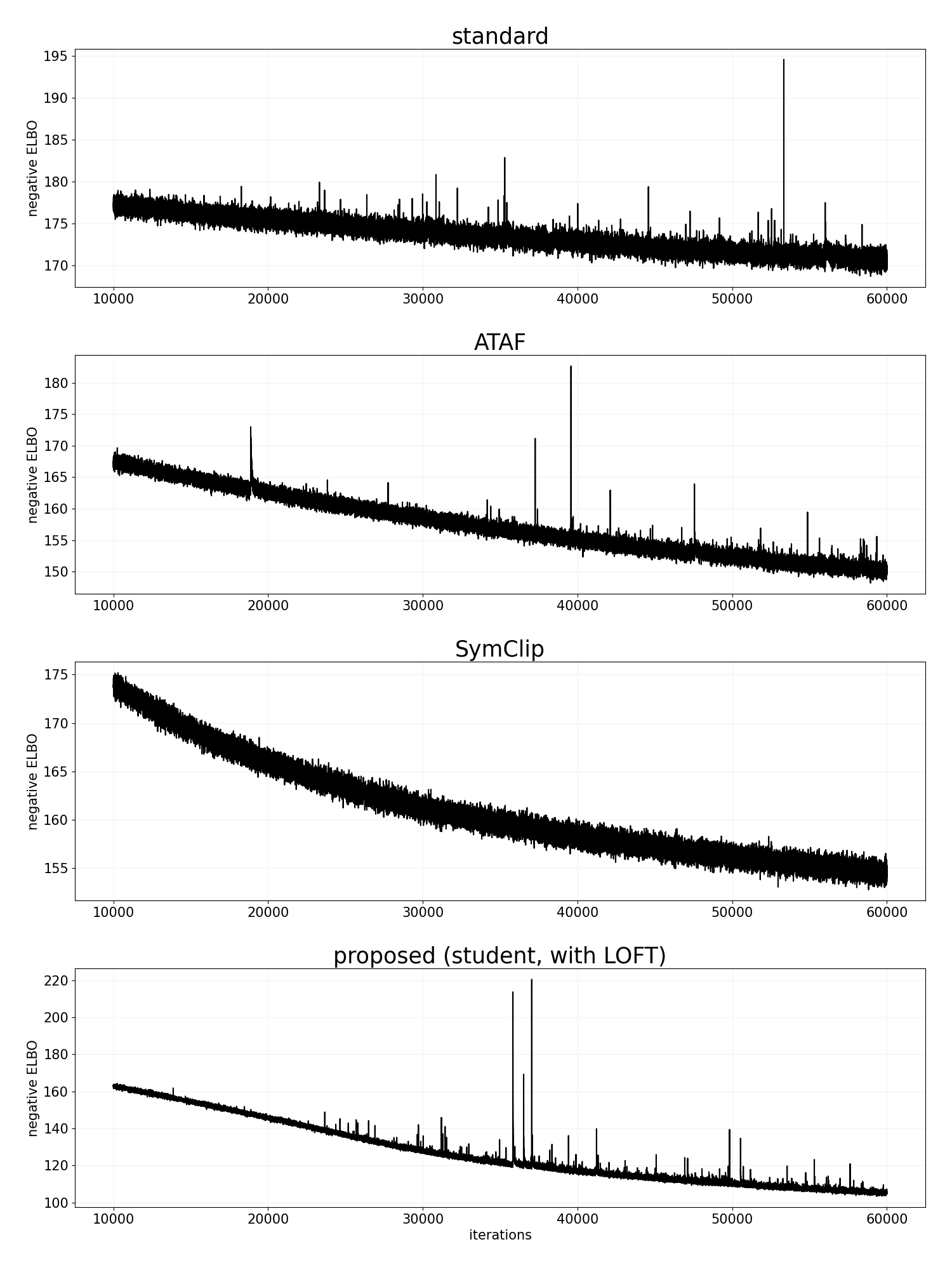}
	\caption{Shows the progress of the loss (= negative ELBO) over the training iterations for the Horseshoe logistic regression model with $d = 2002$.} \label{fig:loss_comparison}
\end{figure*}
\begin{figure*}
	\centering
	\begin{subfigure}{.48\textwidth}
		\centering
		\includegraphics[width=1.0\linewidth,page=1]{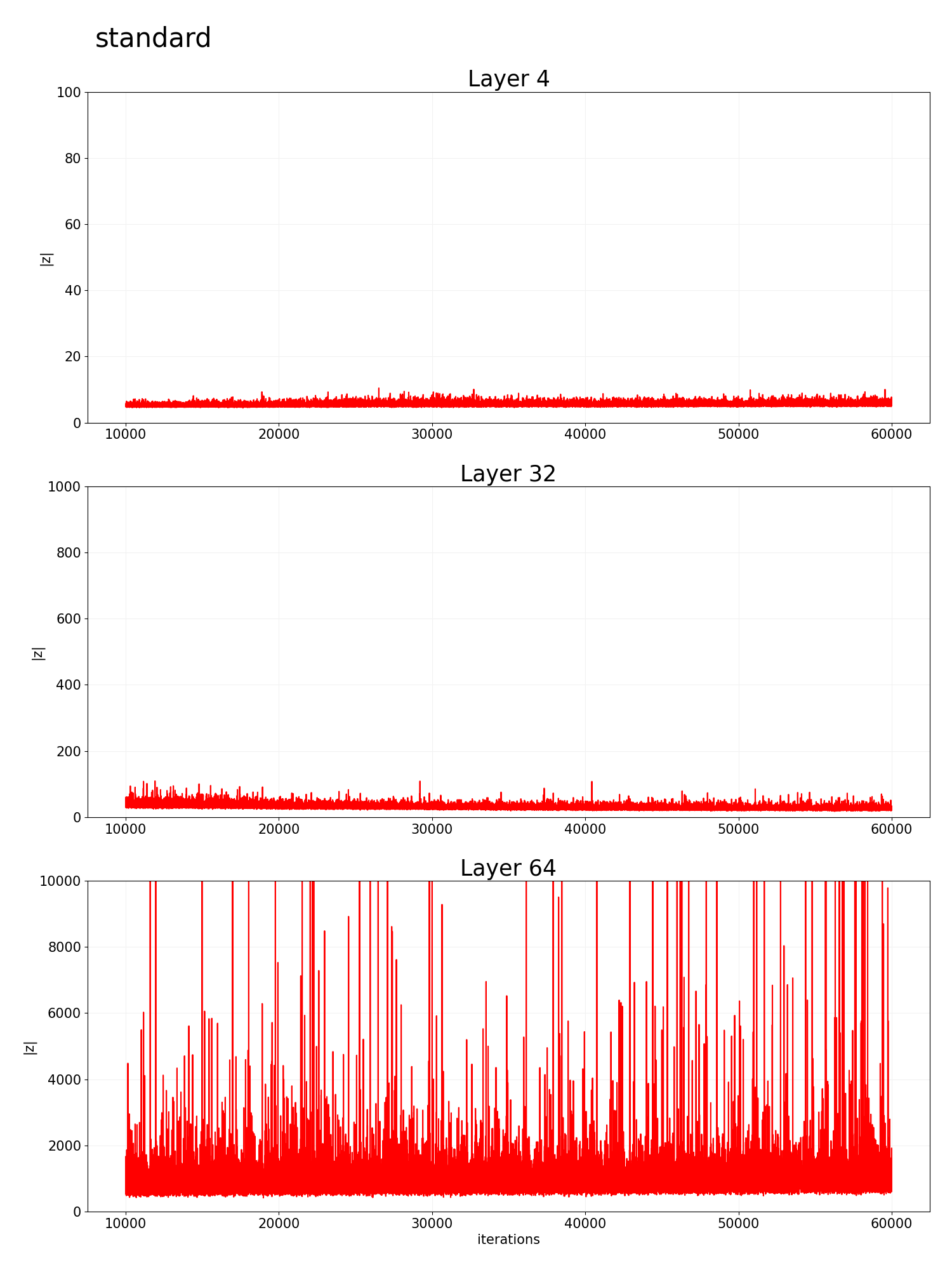}
		\caption{standard}
	\end{subfigure} \quad
	\begin{subfigure}{.48\textwidth}
		\centering
		\includegraphics[width=1.0\linewidth,page=1]{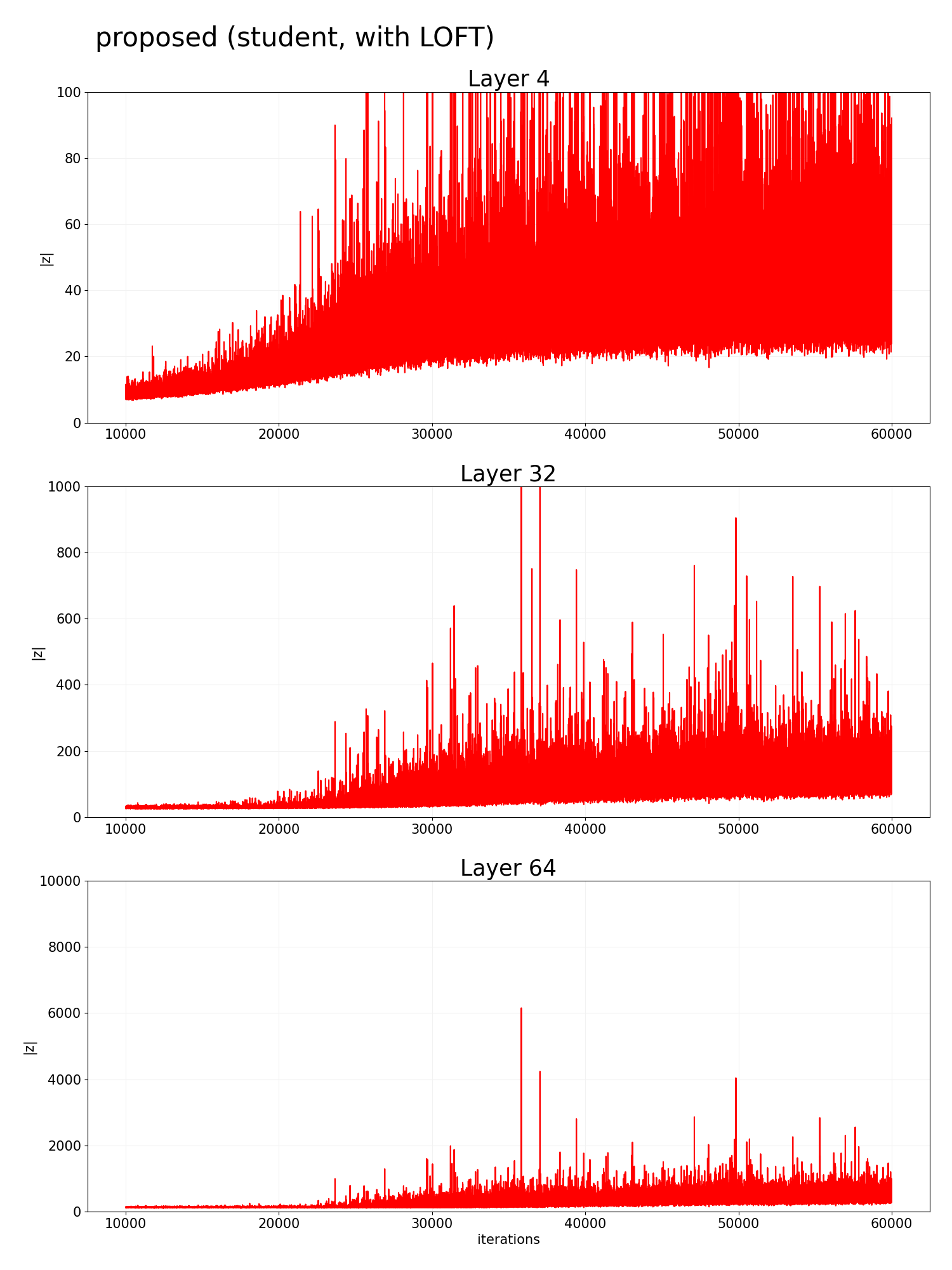}
		\caption{proposed (student with LOFT)}
	\end{subfigure}
	\caption{Shows the maximum value of a mini-batch sample after the 4-th, 32-th, and 64-th layer for the Horseshoe logistic regression model with $d = 2002$ when using (a) the standard normalizing flow, and (b) the proposed method.} \label{fig:layer_samples_HorseshoePriorLogisticRegression}
\end{figure*}
\begin{figure*}
	\centering
	\begin{subfigure}{.48\textwidth}
		\centering
		\includegraphics[width=1.0\linewidth,page=1]{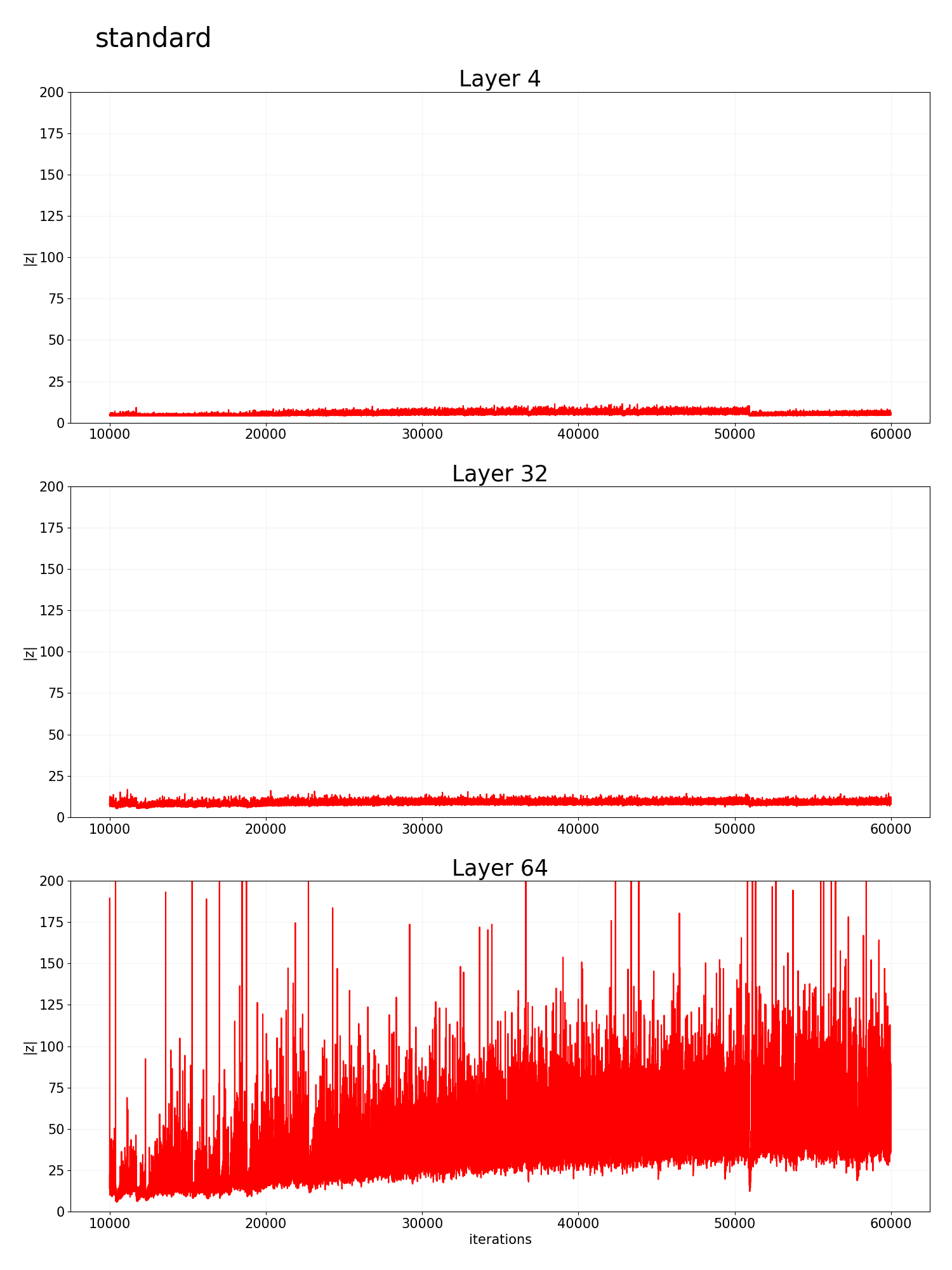}
		\caption{standard}
	\end{subfigure} \quad
	\begin{subfigure}{.48\textwidth}
		\centering
		\includegraphics[width=1.0\linewidth,page=1]{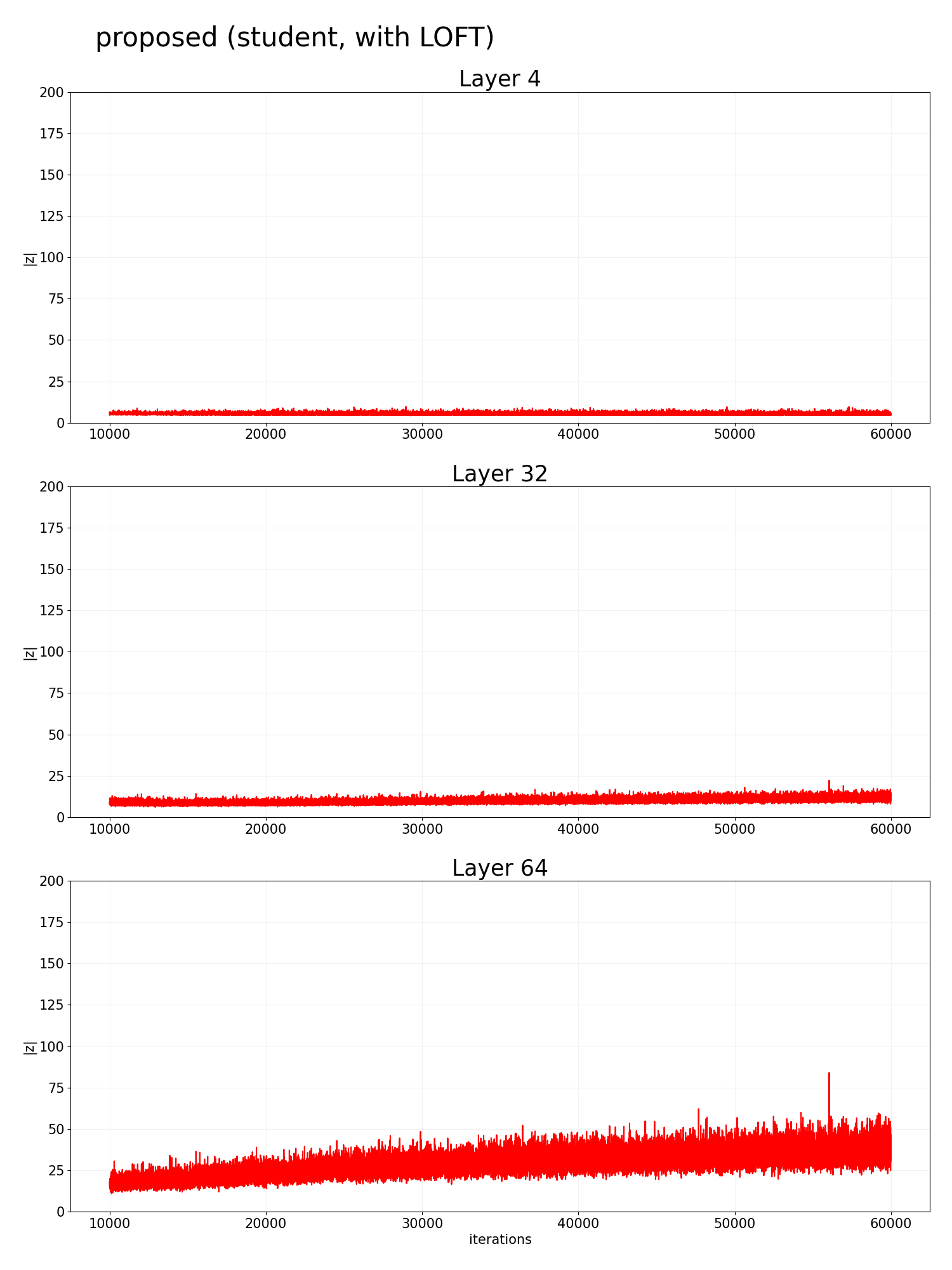}
		\caption{proposed (student with LOFT)}
	\end{subfigure}
	\caption{Shows the maximum value of a mini-batch sample after the 4-th, 32-th, and 64-th layer for the Funnel model with $d = 1000$ when using (a) the standard normalizing flow, and (b) the proposed method.} \label{fig:layer_samples_Funnel}
\end{figure*}

\paragraph{Convergence and Large Sample Values}
First, we take a closer look at the convergence behavior of the proposed method for the Horseshoe logistic regression model.
Figure \ref{fig:loss_comparison} confirms that the proposed method achieves faster convergence to a lower optimum of the loss (= negative ELBO) than other normalizing flow variations.
Next, in Figures \ref{fig:layer_samples_HorseshoePriorLogisticRegression} and \ref{fig:layer_samples_Funnel} we empirically confirm the observation from Section \ref{sec:explodingSampleValues} that normalizing flows based on coupling layers tend to produce larger values with increasing depth. However, comparing the left (a) and right-hand side (b) of the Figures \ref{fig:layer_samples_HorseshoePriorLogisticRegression} and \ref{fig:layer_samples_Funnel} we find that the proposed method successfully restrains large sample values at the last layer.

\paragraph{Important Details of Training Normalizing Flows}
For the training of all normalizing flows, we use double precision (opposite to training of ordinary neural networks, where single precision is often considered sufficient). Furthermore, for training of all normalizing flows we used path gradients (as proposed in \citep{Roeder2017,Vaitl2022}), and do not employ annealing (opposite to the suggestion in \citep{Rezende2015}). 
Our analysis in the appendix shows that these choices are crucial for stable training (see Tables \ref{tab:ELBO_comparison_with_ordinary_KL} and \ref{tab:ELBO_comparison_with_annealing}).\footnote{Moreover we do not use gradient clipping and  l2-regularization as theses had only deteriorate effects.}

The works in \citep{Dhaka2020} show that detecting convergence can be tricky even for variational inference with a Gaussian approximation family.
Indeed, we found that training deep normalizing flows with the ELBO is similar to the mini-batch training of ordinary deep learning: detection of convergence is difficult since after a period of increase, the loss  (negative ELBO) sometimes can drop again. As a consequence, given a fixed budget of $M$ iterations, we compared: (1) taking the model after $M$ iterations, and (2) taking the model with the lowest loss during the iterations $M/2$ and $M$ iterations. Note that at each iteration the loss is evaluate using only 256 samples, whereas for the final evaluation the ELBO is estimated with 20,000 samples.
The results in the Appendix (in Table \ref{tab:ELBO_comparison_with_last_model}) show that the ELBO of the standard Real NVP can be considerably improved by taking the model with the lowest loss, while for the proposed method the differences are negligible.
Note that in favor for the standard model, we used for all of our experiments the lowest loss model.

\paragraph{Importance of Depth of Normalizing Flows}
We also evaluated the influence of the number of coupling layers $r$, i.e. the depth of the normalizing flow.
We compared the ELBO and log marginal likelihoods results for $r = 64$ and $r = 16$, and found that for the proposed method an increase of $r$ consistently improves ELBO and the log marginal likelihood estimate (see Appendix, Figure \ref{tab:ELBO_and_IS_evaluation_horeshoe_synthetic_r64_r16} for the horseshoe model, and \ref{tab:ELBO_comparison_basic_models_r16}, \ref{tab:ELBO_comparison_basic_models_r64}, \ref{tab:IS_comparison_basic_models_r16}, \ref{tab:IS_comparison_basic_models_r64} for all basic models).

 \paragraph{Comparison to HMC sampler}

In the Appendix (in Table \ref{tab:Wassertein_comparison}), we also compare the samples from the normalizing flows to the ground truth, and evaluate in terms of (sliced) Wasserstein distance $\text{WD}(q_{\boldsymbol{\eta}} , p_*)$.
We find that the quality of the samples of the normalizing flows are comparable to the sample of a long run HMC.\footnote{Note that for the Multivariate Student-T model the estimation error of the Wasserstein distance is considerable, making it difficult to state a clear winner.}

\paragraph{Details on Runtime}

A rough estimate of runtimes for all methods are shown in Table \ref{tab:overview_runtimes}.
We measured the runtime using NVIDIA RTX 6000 Ada with 48GB memory, on a Linux server with 32 cores. 
For all variational inference methods we used PyTorch \citep{Paszke2017} and the normflows package \citep{Stimper2023}. 
All implementations, except the HMC from Numpyro \citep{Phan2019}, were using the GPU, as such the runtime for HMC cannot be fairly compared to the other results. All variations of normalizing flows had about the same runtime. 
Note that sampling for normalizing flows is negligible (1$\sim$2 minutes) compared to training.
Therefore, for repeated sampling, normalizing flows are computationally more efficient than SMC and HMC.

\begin{table*}
	\centering
	\caption{Approximate runtimes (wall-clock) for training/sampling of all methods (Horseshoe Logistic Regression model with $d = 2002$, 20 repetitions).} \label{tab:overview_runtimes}
	\footnotesize
	\begin{tabular}{ll}
		\toprule 
		\bfseries Method & \bfseries Runtime in Minutes \\  
		\midrule 
		mean field gaussian &  10 \\   
		proposed (student, with LOFT) &  360 \\   
		SMC &  1510 \\  
		HMC &  33640 \\
		\bottomrule 
	\end{tabular}
\end{table*}

\section{Conclusions and Recommendations} \label{sec:conclusion_and_recommendations}

Our experiments showed that for high dimensional posteriors (more than $1000$ dimensions), normalizing flows based on coupling layers provide better estimates to the marginal likelihood than established methods like SMC. 
For the training objective of all normalizing flows, we used the reverse KL-divergence (ELBO), and verified that better estimates of the ELBO lead to better estimates of the marginal likelihood. 

Furthermore, we found, theoretically and empirically, that normalizing flows based on coupling layers tend to produce samples with high absolute values, whereas the problem becomes more pronounced with increasing depth (=number of coupling layers).
As a remedy, we proposed a soft clipping of the conditional factors $s_i$ and a new log transformation layer named LOFT.
Notably, these modification still preserve the bijectiveness of the flow and allow for computationally efficient calculation of all necessary derivatives.
In particular for high dimensional posteriors with heavy tails we found that our proposed modifications (soft clipping + LOFT) lead to considerably improved estimates of the ELBO and the marginal likelihood.

Finally, we conducted extensive experiments with different modifications in architecture and training, where we found that the following choices lead to good estimates of ELBO/marginal likelihood:
\begin{itemize}
	\item A student-t distribution as base distribution with trainable degree's of freedom for each dimension \citep{Liang2022a}.
	\item A depth of 64 coupling layers (or possibly higher).\footnote{For our experiments, we used 64 coupling layers, and more coupling layers are likely to achieve an even higher ELBO. However, the high memory requirements during back-propagation place an upper bound on the number of coupling layers that can be used in practice.}
	\item Gradient estimate without score term \citep{Roeder2017,Vaitl2022}. 
	\item No annealing of training objective (opposite to suggestion in \citep{Rezende2015}).
	\item Tracking and choosing the model with the smallest loss during training rather than the last model.
\end{itemize}

Though our focus was on the estimation of the marginal likelihood,  our evaluation in terms of Wasserstein distance hints that normalizing flows with the above modifications provide samples with similar quality as HMC. Therefore, combining the above normalizing flows with MCMC as, for example discussed in \citep{Brofos2022,Winter2023}, is a promising direction for future work.

\section*{Funding}
This work was supported by JSPS KAKENHI Grant Number 22K11934.

\section*{Appendix}

\paragraph{Details of Neural Network}

For all normalizing flows, we used for each function $s_i$ and $t_i$ of Equation \eqref{eq:realNVP}, a multi-layer perceptron with one hidden layer of size 100, and ReLU as activation function.
All network parameters were initialized with 0, which means that, before training, output samples equal the distribution of the base distribution.

\paragraph{Details of Synthetic Data}

For the synthetic data, we generated 20 datasets with dimensions $d \in \{10, 100, 1000\}$ and $n = 100$.
The regression data was generated using the linear regression model described in Section 7.2, Example 1 of \citep{tibshirani1996regression}.
The logitistc regression data was generated as follows:
\begin{align*} 
	y_i \stackrel{i.i.d}{\sim} \text{Bernoulli}(\sigma(\mathbf{x}_i^T \boldsymbol{\beta}_0 + \mu_0))  \; \text{, for $i \in \{1, \dots n\}$} \, ,
\end{align*}
with $\boldsymbol{\beta}_0 \in \mathbb{R}^d$, where the first, second, and fifth dimension of $\boldsymbol{\beta}_0$ are set to $3, 1.5$, and $2$, respectively;  $\mu_0 := 1$, and $\mathbf{x}$ are generated from a multivariate normal distribution with mean $\mathbf{0}$ and covariance matrix C, where $C_{ij} := 0.1^{|i-j|}$.

\paragraph{Details of SMC sampler}

For comparison, we use a sequential monte carlo (SMC) sampler that is an extension of annealed importance sampling \citep{Neal2001}, but with improved performance for estimating the normalization constant \citep{arbel2021annealed}. Indeed, removing the resampling step in SMC corresponds to annealed importance sampling \citep{Dai2022}.

For SMC, we used the GPU implementation of \citep{arbel2021annealed} available at \url{https://github.com/google-deepmind/annealed_flow_transport}. As initial distribution $q_0$, we use a standard normal distribution. To bridge between $q_0$ and the target distribution $p_*$, we use a geometric annealing scheme with $100,000$ (intermediate) temperatures.
The number of particles was set to $2000$.\footnote{For Funnel, due to the long runtime, we used only $10,000$ temperatures, and for  $d = 1000$, we had to reduce the number of particles to $100$.}
At each time step (temperature), one iteration of HMC with NUTS is performed. 
Resampling at each time step is performed, if the effective sample size drops below $0.3 N_p$.

\paragraph{Comparison to HMC sampler}

In Table \ref{tab:Wassertein_comparison}, we also evaluate the samples from the normalizing flows to the ground truth, in terms of (sliced) Wasserstein distance $\text{WD}(q_{\boldsymbol{\eta}} , p_*)$, and compare to samples of a long run HMC.
For the HMC we used NUTS \citep{Hoffman2014} with $200,000 \times 4$ chains; discarding first half of each chain as burn-in and thinning with interval 10.

Note that for the Conjugate Linear Regression Model (as described in \ref{sec:targetDistributions_basicModels}) the true posteriors are available as
\begin{align*}
	\log p(\sigma^2 | \mathbf{y}, X) &= \text{Inv-Gamma}(\alpha_{\text{post}}, \beta_{\text{post}}) \, ,  \\
	\log p(\boldsymbol{\beta} | \mathbf{y}, X) &= \text{t}_\nu(\mu_{\text{post}}, \Sigma_{\text{post}}) \, , 
\end{align*}
where $\alpha_{\text{post}} = \frac{1 + n}{2}$,  $\beta_{\text{post}} = \frac{1}{2} ( \| \mathbf{y} \|^2_2 + 1 - \mathbf{y}^T X U^{-1} X^T \mathbf{y})$, $U = X^T X + I$;  $ \text{t}_\nu$ denotes the multivariate student-t distribution with degrees of freedom $\nu = 1 + n$, mean $\mu_{\text{post}} = U^{-1} X^T \mathbf{y}$, and scale matrix $\Sigma_{\text{post}} = \frac{2 \beta}{\nu} U^{-1}$. 

\paragraph{Background on Path Gradients}
Evaluating the gradient at some point $\boldsymbol{\eta}^*$, and using the chain-rule (from matrix calculus), we have
\begin{align*}
	\frac{\partial}{\partial {\boldsymbol{\eta}}}   \log q_{\boldsymbol{\eta}}(f_{\boldsymbol{\eta}}(\mathbf{z})) \Big|_{\boldsymbol{\eta} = \boldsymbol{\eta}^*}  = 
	\Big( \frac{\partial}{\partial {\boldsymbol{\theta}}}   \log q_{\boldsymbol{\eta}^*}(\boldsymbol{\theta}) \Big) \frac{\partial}{\partial {\boldsymbol{\eta}}} 
	f_{\boldsymbol{\eta}}(\mathbf{z}) + \frac{\partial}{\partial {\boldsymbol{\eta}}} 
	\log q_{\boldsymbol{\eta}}(\boldsymbol{\theta})\Big|_{\boldsymbol{\theta} = f_{\boldsymbol{\eta}^*}(\mathbf{z})}
\end{align*}
and
\begin{align*}
	\frac{\partial}{\partial {\boldsymbol{\eta}}}  \log p(f_{\boldsymbol{\eta}}(\mathbf{z}), D)  &= \frac{\partial}{\partial {\boldsymbol{\eta}}}  \log p(f_{\boldsymbol{\eta}}(\mathbf{z}) | D)  \\
	&= \Big( \frac{\partial}{\partial {\boldsymbol{\theta}}}   \log p(\boldsymbol{\theta} | D) \Big) \frac{\partial}{\partial {\boldsymbol{\eta}}} 
	f_{\boldsymbol{\eta}}(\mathbf{z}) \, .
\end{align*}
Putting this together, we get
\begin{align*} 
	\frac{\partial}{\partial {\boldsymbol{\eta}}} \text{KL}(q_{\boldsymbol{\eta}} || p_*) 
	&=  \mathbb{E}_{q_0} \Big[   \frac{\partial}{\partial {\boldsymbol{\eta}}} \log \Big( \frac{q_{\boldsymbol{\eta}}(f_{\boldsymbol{\eta}}(\mathbf{z}))}{p(f_{\boldsymbol{\eta}}(\mathbf{z}), D)} \Big) \Big] \\
	&=  \mathbb{E}_{q_0} \Big[  \Big( \frac{\partial}{\partial {\boldsymbol{\theta}}}   \log q_{\boldsymbol{\eta}^*}(\boldsymbol{\theta}) \Big) \frac{\partial}{\partial {\boldsymbol{\eta}}} 
	f_{\boldsymbol{\eta}}(\mathbf{z}) + \frac{\partial}{\partial {\boldsymbol{\eta}}} 
	\log q_{\boldsymbol{\eta}}(\boldsymbol{\theta})\Big|_{\boldsymbol{\theta} = f_{\boldsymbol{\eta}^*}(\mathbf{z})} - \Big( \frac{\partial}{\partial {\boldsymbol{\theta}}}  \log p(\boldsymbol{\theta} | D) \Big) \frac{\partial}{\partial {\boldsymbol{\eta}}} 
	f_{\boldsymbol{\eta}}(\mathbf{z})  \Big]  \\
	&=  \mathbb{E}_{q_0} \Big[  \Big( \frac{\partial}{\partial {\boldsymbol{\theta}}}   \log q_{\boldsymbol{\eta}^*}(\boldsymbol{\theta})
	-  \frac{\partial}{\partial {\boldsymbol{\theta}}}  \log p(\boldsymbol{\theta} | D) \Big)  \frac{\partial}{\partial {\boldsymbol{\eta}}}  
	f_{\boldsymbol{\eta}}(\mathbf{z}) 
	+ \frac{\partial}{\partial {\boldsymbol{\eta}}} 
	\log q_{\boldsymbol{\eta}}(\boldsymbol{\theta})\Big|_{\boldsymbol{\theta} = f_{\boldsymbol{\eta}^*}(\mathbf{z})} \Big] \, .
\end{align*}
Assuming that $q_{\boldsymbol{\eta}^*}(\boldsymbol{\theta}) \approx p(\boldsymbol{\theta} | D)$, we get 
\begin{align*} 
	\frac{\partial}{\partial {\boldsymbol{\eta}}} \text{KL}(q_{\boldsymbol{\eta}} || p_*)
	&\approx  \mathbb{E}_{q_0} \Big[  \frac{\partial}{\partial {\boldsymbol{\eta}}}
	\log q_{\boldsymbol{\eta}}(\boldsymbol{\theta})\Big|_{\boldsymbol{\theta} = f_{\boldsymbol{\eta}^*}(\mathbf{z})} \Big] \, , 
\end{align*}
whereas for samples $\mathbf{z}_k \sim q_0$, $k = 1,\ldots,b$,  the estimator is 
\begin{align*}
	\frac{1}{b} \sum_{k = 1}^b \Big[  \frac{\partial}{\partial {\boldsymbol{\eta}}} 
	\log q_{\boldsymbol{\eta}}(\boldsymbol{\theta})\Big|_{\boldsymbol{\theta} = f_{\boldsymbol{\eta}^*}(\mathbf{z}_k)} \Big] \, .
\end{align*}
Note that the variance of this estimator is not $0$. 
Therefore, the works in \citep{Roeder2017,Vaitl2022} propose to use instead the estimator 
\begin{align*} 
	\frac{1}{b} \sum_{k = 1}^b \Big[  \frac{\partial}{\partial {\boldsymbol{\eta}}} \log \Big( \frac{q_{\boldsymbol{\eta}}(f_{\boldsymbol{\eta}}(\mathbf{z}_k))}{p(f_{\boldsymbol{\eta}}(\mathbf{z}_k), D)} \Big) -  \frac{\partial}{\partial {\boldsymbol{\eta}}} 
	\log q_{\boldsymbol{\eta}}(\boldsymbol{\theta})\Big|_{\boldsymbol{\theta} = f_{\boldsymbol{\eta}^*}(\mathbf{z}_k)}  \Big] \, , 
\end{align*}
which has variance 0 when $q_{\boldsymbol{\eta}^*}(\boldsymbol{\theta}) = p(\boldsymbol{\theta} | D)$.

\begin{table*}
	\centering
	\caption{Evaluation of all methods in terms of Wasserstein distance to true distribution (standard deviation in brackets) for different dimensions $d$. For estimation the sliced Wasserstein distance is used; "nan" means that a numerically stable estimate could not be found; number of flows $r = 64$.} \label{tab:Wassertein_comparison}
	\footnotesize
	\begin{tabular}{llll}
		\toprule 
		\multicolumn{4}{c}{  Funnel  } \\ 
		\midrule 
		\bfseries Method & \bfseries $d = 10$ & \bfseries $d = 100$ & \bfseries $d = 1000$ \\  
		\midrule 
		mean field gaussian &  7.96075 (0.5571) &  9.16279 (1.95161) &  8.81864 (0.76501) \\  
		standard &  \textbf{3.55096} (0.52902) &  5.49619 (0.75929) &  7.05587 (1.12632) \\  
		ATAF &  6.16254 (1.27847) &  5.43344 (1.35165) &  7.08017 (0.71492) \\  
		SymClip &  5.464 (0.79736) &  6.39209 (0.76325) &  7.35277 (0.88327) \\  
		proposed (gaussian, with LOFT) &  5.26478 (1.5853) &  7.9078 (5.04944) &  7.17303 (0.69758) \\  
		proposed (student, with LOFT) &  5.10961 (0.97158) &  6.91015 (1.97344) &  7.28766 (0.68416) \\  
		proposed (gaussian, no LOFT) &  4.85281 (1.25448) &  6.04271 (0.88001) &  7.15134 (0.75477) \\  
		proposed (student, no LOFT) &  5.30764 (1.43497) &  6.48703 (1.31521) &  7.50812 (1.26615) \\  
		HMC &  3.80284 (1.06943) &  \textbf{3.07891} (1.14817) &  \textbf{6.23767} (2.1133) \\
		\midrule 
		\multicolumn{4}{c}{  Multivariate Student-T  } \\ 
		\midrule 
		\bfseries Method & \bfseries $d = 10$ & \bfseries $d = 100$ & \bfseries $d = 1000$ \\  
		\midrule 
		mean field gaussian &  978.22757 (2077.58198) &  947.55754 (1981.72515) &  \textbf{339.85119} (319.99747) \\  
		standard &  \textbf{413.43534} (377.51084) &  643.26898 (950.03763) &  491.19396 (410.11727) \\  
		ATAF &  813.73265 (1054.58138) &  434.39064 (487.75902) &  nan (nan) \\  
		SymClip &  908.83123 (1339.8407) &  813.47494 (1260.90776) &  1675.82688 (3798.4308) \\  
		proposed (gaussian, with LOFT) &  421.94993 (448.44594) &  431.86027 (502.31684) &  525.45636 (816.70971) \\  
		proposed (student, with LOFT) &  1087.58516 (2393.59627) &  465.30794 (626.16168) &  581.85254 (789.12662) \\  
		proposed (gaussian, no LOFT) &  424.74731 (537.22081) &  368.34586 (307.89202) &  1073.10593 (3099.12445) \\  
		proposed (student, no LOFT) &  661.30146 (1076.37543) &  \textbf{235.95782} (153.88255) &  630.76724 (1291.59487) \\  
		HMC &  1319.2934 (1812.88423) &  840.564 (622.86178) &  788.89202 (1167.01651) \\
		\midrule 
		\multicolumn{4}{c}{  Multivariate Gaussian Mixture  } \\ 
		\midrule 
		\bfseries Method & \bfseries $d = 10$ & \bfseries $d = 100$ & \bfseries $d = 1000$ \\  
		\midrule 
		mean field gaussian &  1.00168 (0.00169) &  0.16096 (0.00062) &  0.02499 (0.00015) \\  
		standard &  0.18057 (0.00635) &  0.02511 (0.00172) &  \textbf{0.01405} (0.00015) \\  
		ATAF &  0.12186 (0.00771) &  0.04374 (0.00215) &  0.02392 (0.00075) \\  
		SymClip &  0.14645 (0.00652) &  0.04968 (0.00214) &  0.01622 (0.00047) \\  
		proposed (gaussian, with LOFT) &  0.03911 (0.00337) &  0.01673 (0.00054) &  0.02191 (0.00065) \\  
		proposed (student, with LOFT) &  0.20198 (0.00806) &  \textbf{0.01652} (0.00044) &  0.01801 (0.00054) \\  
		proposed (gaussian, no LOFT) &  0.03906 (0.00351) &  0.01654 (0.00033) &  0.02195 (0.00064) \\  
		proposed (student, no LOFT) &  0.147 (0.00912) &  0.05994 (0.00206) &  0.01686 (0.00048) \\  
		HMC &  \textbf{0.0257} (0.00654) &  0.01947 (0.00155) &  0.01597 (0.00049) \\
		\midrule 
		\multicolumn{4}{c}{  Conjugate Linear Regression  } \\ 
		\midrule 
		\bfseries Method & \bfseries $d = 11$ & \bfseries $d = 101$ & \bfseries $d = 1001$ \\  
		\midrule 
		mean field gaussian &  0.15449 (0.00047) &  0.31762 (0.00033) &  0.12625 (0.00006) \\  
		standard &  0.00787 (0.00045) &  0.00693 (0.00024) &  0.00972 (0.00008) \\  
		ATAF &  0.00648 (0.00033) &  0.00712 (0.00025) &  0.00545 (0.00005) \\  
		SymClip &  0.00603 (0.00027) &  0.00679 (0.00015) &  0.00591 (0.00005) \\  
		proposed (gaussian, with LOFT) &  0.00723 (0.00042) &  0.00658 (0.00024) &  0.00635 (0.00004) \\  
		proposed (student, with LOFT) &  0.00646 (0.00034) &  0.00686 (0.0002) &  0.00562 (0.00004) \\  
		proposed (gaussian, no LOFT) &  0.00694 (0.00034) &  0.00664 (0.00021) &  0.00636 (0.00003) \\  
		proposed (student, no LOFT) &  0.0066 (0.0004) &  0.0069 (0.00022) &  0.00561 (0.00004) \\  
		HMC &  \textbf{0.00521} (0.00036) &  \textbf{0.00595} (0.00025) &  \textbf{0.00259} (0.00002) \\
		\bottomrule 
	\end{tabular}
\end{table*}

\begin{table*}
	\centering
	\caption{Comparison of ELBO results with and without annealing for $d = 1000$ (for the Conjugate Linear Regression model $d = 1001$). "nan" means that a numerically stable estimate could not be found} \label{tab:ELBO_comparison_with_annealing}
	\footnotesize
	\begin{tabular}{lll}
		\toprule 
		\multicolumn{3}{c}{  Funnel  } \\ 
		\midrule 
		\bfseries Method & without annealing & with annealing \\  
		\midrule 
		mean field gaussian &  \textbf{-4.20619} (0.00786) &  -4.25079 (0.00797) \\  
		standard &  \textbf{-0.13076} (0.00262) &  -0.71427 (0.00653) \\  
		ATAF &  \textbf{-0.19676} (0.00382) &  -11.87143 (0.04698) \\  
		SymClip &  \textbf{-0.19903} (0.0034) &  -0.26312 (0.00427) \\  
		proposed (gaussian, with LOFT) &  -0.16222 (0.00316) &  \textbf{-0.15195} (0.00414) \\  
		proposed (student, with LOFT) &  \textbf{-0.16584} (0.0034) &  -0.20455 (0.00309) \\
		\midrule 
		\multicolumn{3}{c}{  Multivariate Student-T  } \\ 
		\midrule 
		\bfseries Method & without annealing & with annealing \\  
		\midrule 
		mean field gaussian &  \textbf{-7.35684} (0.00516) &  -7.56552 (0.00736) \\  
		standard &  \textbf{-1.07369} (0.06991) &  nan (nan) \\  
		ATAF &  \textbf{-1.51959} (0.44861) &  nan (nan) \\  
		SymClip &  \textbf{-0.33732} (0.0055) &  -0.36437 (0.00487) \\  
		proposed (gaussian, with LOFT) &  \textbf{-0.23245} (0.00392) &  -0.39716 (0.00437) \\  
		proposed (student, with LOFT) &  \textbf{-0.29187} (0.00531) &  -0.52886 (0.00652) \\
		\midrule 
		\multicolumn{3}{c}{  Multivariate Gaussian Mixture  } \\ 
		\midrule 
		\bfseries Method & without annealing & with annealing \\  
		\midrule 
		mean field gaussian &  -1.09121 (0.00081) &  \textbf{-1.091} (0.00089) \\  
		standard &  \textbf{-0.14578} (0.0044) &  -1.16493 (0.00271) \\  
		ATAF &  \textbf{-0.19254} (0.00561) &  -1.35144 (0.01046) \\  
		SymClip &  -0.17354 (0.00437) &  \textbf{-0.1642} (0.00308) \\  
		proposed (gaussian, with LOFT) &  \textbf{-0.15943} (0.00418) &  -0.1924 (0.00507) \\  
		proposed (student, with LOFT) &  -0.20653 (0.00458) &  \textbf{-0.18603} (0.00446) \\
		\midrule 
		\multicolumn{3}{c}{  Conjugate Linear Regression  } \\ 
		\midrule 
		\bfseries Method & without annealing & with annealing \\  
		\midrule 
		mean field gaussian &  -2348.47948 (0.61318) &  \textbf{-2339.86726} (0.45202) \\  
		standard &  \textbf{-327.26313} (0.0318) &  nan (nan) \\  
		ATAF &  \textbf{-328.46092} (0.0203) &  nan (nan) \\  
		SymClip &  \textbf{-327.61051} (0.02208) &  -27497.59608 (25.57452) \\  
		proposed (gaussian, with LOFT) &  \textbf{-330.37689} (0.02949) &  -334.25809 (0.03358) \\  
		proposed (student, with LOFT) &  \textbf{-330.41806} (0.0309) &  -333.41908 (0.04207) \\
		\bottomrule 
	\end{tabular}
\end{table*}

\begin{table*}
	\centering
	\caption{Comparison of ELBO results with and without score term for $d = 1000$ (for the Conjugate Linear Regression model $d = 1001$). } \label{tab:ELBO_comparison_with_ordinary_KL}
	\footnotesize
	\begin{tabular}{lll}
		\toprule 
\multicolumn{3}{c}{  Funnel  } \\ 
\midrule 
\bfseries Method & without score term & with score term \\  
\midrule 
mean field gaussian &  \textbf{-4.20619} (0.00786) &  -4.21122 (0.00646) \\  
standard &  \textbf{-0.13076} (0.00262) &  -1.23624 (0.00825) \\  
ATAF &  \textbf{-0.19676} (0.00382) &  -1.34531 (0.00832) \\  
SymClip &  \textbf{-0.19903} (0.0034) &  -0.84768 (0.00697) \\  
proposed (gaussian, with LOFT) &  \textbf{-0.16222} (0.00316) &  -0.82072 (0.0075) \\  
proposed (student, with LOFT) &  \textbf{-0.16584} (0.0034) &  -0.93281 (0.00817) \\
\midrule 
\multicolumn{3}{c}{  Multivariate Student-T  } \\ 
\midrule 
\bfseries Method & without score term & with score term \\  
\midrule 
mean field gaussian &  \textbf{-7.35684} (0.00516) &  -7.63274 (0.00774) \\  
standard &  \textbf{-1.07369} (0.06991) &  -3.67112 (0.0079) \\  
ATAF &  \textbf{-1.51959} (0.44861) &  -3.71659 (0.01131) \\  
SymClip &  \textbf{-0.33732} (0.0055) &  -3.59259 (0.01113) \\  
proposed (gaussian, with LOFT) &  \textbf{-0.23245} (0.00392) &  -3.55175 (0.00922) \\  
proposed (student, with LOFT) &  \textbf{-0.29187} (0.00531) &  -3.59585 (0.00792) \\
\midrule 
\multicolumn{3}{c}{  Multivariate Gaussian Mixture  } \\ 
\midrule 
\bfseries Method & without score term & with score term \\  
\midrule 
mean field gaussian &  \textbf{-1.09121} (0.00081) &  -1.09444 (0.00111) \\  
standard &  \textbf{-0.14578} (0.0044) &  -1.22924 (0.00413) \\  
ATAF &  \textbf{-0.19254} (0.00561) &  -1.27836 (0.00491) \\  
SymClip &  \textbf{-0.17354} (0.00437) &  -1.20088 (0.00392) \\  
proposed (gaussian, with LOFT) &  \textbf{-0.15943} (0.00418) &  -1.20163 (0.00347) \\  
proposed (student, with LOFT) &  \textbf{-0.20653} (0.00458) &  -1.24067 (0.0039) \\
\midrule 
\multicolumn{3}{c}{  Conjugate Linear Regression  } \\ 
\midrule 
\bfseries Method & without score term & with score term \\  
\midrule 
mean field gaussian &  -2348.47948 (0.61318) &  \textbf{-2290.86055} (0.53819) \\  
standard &  \textbf{-327.26313} (0.0318) &  -336.70439 (0.04761) \\  
ATAF &  \textbf{-328.46092} (0.0203) &  -335.94146 (0.03875) \\  
SymClip &  \textbf{-327.61051} (0.02208) &  -335.4258 (0.03106) \\  
proposed (gaussian, with LOFT) &  \textbf{-330.37689} (0.02949) &  -336.43568 (0.03155) \\  
proposed (student, with LOFT) &  \textbf{-330.41806} (0.0309) &  -336.42227 (0.03385) \\
		\bottomrule 
	\end{tabular}
\end{table*}

\begin{table*}
	\centering
	\caption{Comparison of ELBO results using either the model with the lowest loss during training, or the last model for $d = 1000$ (for the Conjugate Linear Regression model $d = 1001$). } \label{tab:ELBO_comparison_with_last_model} 
	\footnotesize
	\begin{tabular}{lll}
		\toprule 
		\multicolumn{3}{c}{  Funnel  } \\ 
		\midrule 
		\bfseries Method & lowest loss model & last model \\  
		\midrule 
		mean field gaussian &  -4.20619 (0.00786) &  \textbf{-4.20447} (0.00711) \\  
		standard &  \textbf{-0.13076} (0.00262) &  -0.17381 (0.00354) \\  
		ATAF &  \textbf{-0.19676} (0.00382) &  -0.2487 (0.00584) \\  
		SymClip &  -0.19903 (0.0034) &  \textbf{-0.17057} (0.004) \\  
		proposed (gaussian, with LOFT) &  \textbf{-0.16222} (0.00316) &  -0.17952 (0.00274) \\  
		proposed (student, with LOFT) &  -0.16584 (0.0034) &  \textbf{-0.15643} (0.00314) \\
		\midrule 
		\multicolumn{3}{c}{  Multivariate Student-T  } \\ 
		\midrule 
		\bfseries Method & lowest loss model & last model \\  
		\midrule 
		mean field gaussian &  -7.35684 (0.00516) &  \textbf{-7.32763} (0.0068) \\  
		standard &  \textbf{-1.07369} (0.06991) &  -1.40608 (0.03054) \\  
		ATAF &  \textbf{-1.51959} (0.44861) &  -5.34154 (0.02333) \\  
		SymClip &  -0.33732 (0.0055) &  \textbf{-0.29835} (0.00542) \\  
		proposed (gaussian, with LOFT) &  \textbf{-0.23245} (0.00392) &  -0.25133 (0.00541) \\  
		proposed (student, with LOFT) &  -0.29187 (0.00531) &  \textbf{-0.27859} (0.0046) \\
		\midrule 
		\multicolumn{3}{c}{  Multivariate Gaussian Mixture  } \\ 
		\midrule 
		\bfseries Method & lowest loss model & last model \\  
		\midrule 
		mean field gaussian &  -1.09121 (0.00081) &  \textbf{-1.09118} (0.00091) \\  
		standard &  \textbf{-0.14578} (0.0044) &  -0.23477 (0.00475) \\  
		ATAF &  -0.19254 (0.00561) &  \textbf{-0.18509} (0.00486) \\  
		SymClip &  -0.17354 (0.00437) &  \textbf{-0.15595} (0.00344) \\  
		proposed (gaussian, with LOFT) &  \textbf{-0.15943} (0.00418) &  -0.1871 (0.00403) \\  
		proposed (student, with LOFT) &  -0.20653 (0.00458) &  \textbf{-0.19927} (0.00413) \\
		\midrule 
		\multicolumn{3}{c}{  Conjugate Linear Regression  } \\ 
		\midrule 
		\bfseries Method & lowest loss model & last model \\  
		\midrule 
		mean field gaussian &  -2348.47948 (0.61318) &  \textbf{-2347.61854} (0.47442) \\  
		standard &  \textbf{-327.26313} (0.0318) &  -329.39336 (0.02498) \\  
		ATAF &  \textbf{-328.46092} (0.0203) &  -332.94404 (0.03647) \\  
		SymClip &  \textbf{-327.61051} (0.02208) &  -331.89927 (0.04517) \\  
		proposed (gaussian, with LOFT) &  \textbf{-330.37689} (0.02949) &  -331.51712 (0.03626) \\  
		proposed (student, with LOFT) &  \textbf{-330.41806} (0.0309) &  -331.7316 (0.03451) \\
		\bottomrule 
	\end{tabular}
\end{table*}


\begin{table*}
	\centering
	\caption{Evaluation of all methods in terms of ELBO (standard deviation in brackets) for $d \in \{10, 100, 1000\}$; 
		number of flows $r = 64$.} \label{tab:ELBO_comparison_basic_models_r64}
	\footnotesize
	\begin{tabular}{llll}
		\toprule 
		\multicolumn{4}{c}{  Funnel  } \\ 
		\midrule 
		\bfseries Method & \bfseries $d = 10$ & \bfseries $d = 100$ & \bfseries $d = 1000$ \\  
		\midrule 
		mean field gaussian &  -1.86318 (0.00707) &  -3.0504 (0.00729) &  -4.20619 (0.00786) \\  
		standard &  -0.01344 (0.00117) &  -0.0566 (0.00193) &  \textbf{-0.13076} (0.00262) \\  
		ATAF &  -0.05574 (0.00697) &  -0.05184 (0.00165) &  -0.19676 (0.00382) \\  
		SymClip &  -0.02905 (0.00114) &  -0.07004 (0.00131) &  -0.19903 (0.0034) \\  
		proposed (gaussian, with LOFT) &  \textbf{-0.00896} (0.00066) &  -0.04321 (0.00187) &  -0.16222 (0.00316) \\  
		proposed (student, with LOFT) &  -0.01244 (0.00072) &  -0.0404 (0.00153) &  -0.16584 (0.0034) \\  
		proposed (gaussian, no LOFT) &  -0.01111 (0.00084) &  -0.04291 (0.00156) &  -0.16612 (0.0029) \\  
		proposed (student, no LOFT) &  -0.01318 (0.00082) &  \textbf{-0.0402} (0.00154) &  -0.16584 (0.0034) \\
		\midrule 
		\multicolumn{4}{c}{  Multivariate Student-T  } \\ 
		\midrule 
		\bfseries Method & \bfseries $d = 10$ & \bfseries $d = 100$ & \bfseries $d = 1000$ \\  
		\midrule 
		mean field gaussian &  -2.16601 (0.00651) &  -4.5299 (0.00767) &  -7.35684 (0.00516) \\  
		standard &  -0.02437 (0.00138) &  -0.08749 (0.00483) &  -1.07369 (0.06991) \\  
		ATAF &  -0.03101 (0.00173) &  -0.10369 (0.0026) &  -1.51959 (0.44861) \\  
		SymClip &  -0.01674 (0.00075) &  -0.05244 (0.00131) &  -0.33732 (0.0055) \\  
		proposed (gaussian, with LOFT) &  -0.01722 (0.00052) &  \textbf{-0.04715} (0.00146) &  \textbf{-0.23245} (0.00392) \\  
		proposed (student, with LOFT) &  -0.01815 (0.00102) &  -0.06186 (0.00189) &  -0.29187 (0.00531) \\  
		proposed (gaussian, no LOFT) &  -0.0164 (0.00061) &  -0.05737 (0.00199) &  -0.2555 (0.00421) \\  
		proposed (student, no LOFT) &  \textbf{-0.0158} (0.00077) &  -0.05686 (0.00217) &  -0.27904 (0.00403) \\
		\midrule 
		\multicolumn{4}{c}{  Multivariate Gaussian Mixture  } \\ 
		\midrule 
		\bfseries Method & \bfseries $d = 10$ & \bfseries $d = 100$ & \bfseries $d = 1000$ \\  
		\midrule 
		mean field gaussian &  -1.09054 (0.00072) &  -1.09085 (0.00086) &  -1.09121 (0.00081) \\  
		standard &  -0.01622 (0.00128) &  \textbf{-0.04753} (0.00198) &  \textbf{-0.14578} (0.0044) \\  
		ATAF &  -0.01508 (0.00118) &  -0.08112 (0.00245) &  -0.19254 (0.00561) \\  
		SymClip &  -0.01116 (0.00095) &  -0.05408 (0.0024) &  -0.17354 (0.00437) \\  
		proposed (gaussian, with LOFT) &  \textbf{-0.00966} (0.00098) &  -0.06171 (0.0025) &  -0.15943 (0.00418) \\  
		proposed (student, with LOFT) &  -0.01534 (0.00108) &  -0.06713 (0.00258) &  -0.20653 (0.00458) \\  
		proposed (gaussian, no LOFT) &  -0.00966 (0.00098) &  -0.06171 (0.0025) &  -0.15943 (0.00418) \\  
		proposed (student, no LOFT) &  -0.01138 (0.00105) &  -0.07312 (0.00248) &  -0.19908 (0.00292) \\
		\midrule 
		\multicolumn{4}{c}{  Conjugate Linear Regression  } \\ 
		\midrule 
		\bfseries Method & \bfseries $d = 11$ & \bfseries $d = 101$ & \bfseries $d = 1001$ \\  
		\midrule 
		mean field gaussian &  -275.82849 (0.0274) &  -399.32425 (0.05422) &  -2348.47948 (0.61318) \\  
		standard &  -269.74157 (0.00046) &  -347.86269 (0.00181) &  \textbf{-327.26313} (0.0318) \\  
		ATAF &  -269.74116 (0.00032) &  -347.86356 (0.0016) &  -328.46092 (0.0203) \\  
		SymClip &  \textbf{-269.74059} (0.00035) &  -347.87454 (0.00245) &  -327.61051 (0.02208) \\  
		proposed (gaussian, with LOFT) &  -269.74193 (0.0005) &  \textbf{-347.84601} (0.00193) &  -330.37689 (0.02949) \\  
		proposed (student, with LOFT) &  -269.74097 (0.00039) &  -347.8542 (0.00208) &  -330.41806 (0.0309) \\  
		proposed (gaussian, no LOFT) &  -269.74193 (0.0005) &  -347.84601 (0.00193) &  -330.37689 (0.02949) \\  
		proposed (student, no LOFT) &  -269.74097 (0.00039) &  -347.8542 (0.00208) &  -330.41806 (0.0309) \\
		\bottomrule 
	\end{tabular}
\end{table*}

\begin{table*}
	\centering
	\caption{Estimated log marginal likelihood (standard deviation in brackets) for $d \in \{10, 100, 1000\}$; 
		number of flows $r = 64$.} \label{tab:IS_comparison_basic_models_r64}
	\footnotesize
	\begin{tabular}{llll}
		\toprule 
		\multicolumn{4}{c}{  Funnel  } \\ 
		\midrule 
		\bfseries Method & \bfseries $d = 10$ & \bfseries $d = 100$ & \bfseries $d = 1000$ \\  
		\midrule 
		true value & 0.0 & 0.0 & 0.0  \\ 
		\midrule
		mean field gaussian &  -0.9051 (0.25602) &  -1.93821 (0.67967) &  -3.11768 (0.52003) \\  
		standard &  \textbf{-0.00068} (0.00282) &  -0.00615 (0.00308) &  \textbf{-0.02001} (0.01194) \\  
		ATAF &  -0.00992 (0.01665) &  \textbf{0.00199} (0.01963) &  -0.02826 (0.01323) \\  
		SymClip &  -0.00545 (0.0103) &  -0.03279 (0.00632) &  -0.06983 (0.01259) \\  
		proposed (gaussian, with LOFT) &  -0.00184 (0.00381) &  -0.01278 (0.0038) &  -0.03778 (0.02408) \\  
		proposed (student, with LOFT) &  -0.00267 (0.00598) &  -0.01402 (0.00323) &  -0.04245 (0.00766) \\  
		proposed (gaussian, no LOFT) &  -0.00129 (0.00292) &  -0.01385 (0.00356) &  -0.04302 (0.01002) \\  
		proposed (student, no LOFT) &  -0.00283 (0.00468) &  -0.01393 (0.00334) &  -0.04245 (0.00766) \\  
		SMC &  -0.1429 (0.06911) &  -1.76495 (0.36939) &  -3.50737 (0.46121) \\
		\midrule 
		\multicolumn{4}{c}{  Multivariate Student-T  } \\ 
		\midrule 
		\bfseries Method & \bfseries $d = 10$ & \bfseries $d = 100$ & \bfseries $d = 1000$ \\  
		\midrule 
		true value & 0.0 & 0.0 & 0.0  \\ 
		\midrule
		mean field gaussian &  -0.92496 (0.16171) &  -2.87842 (0.3654) &  -5.89942 (0.22141) \\  
		standard &  -0.00642 (0.00224) &  -0.01491 (0.00444) &  -0.19315 (0.04714) \\  
		ATAF &  \textbf{-0.00416} (0.01041) &  -0.0157 (0.00605) &  -0.10957 (0.09367) \\  
		SymClip &  -0.01059 (0.00149) &  \textbf{-0.00923} (0.03558) &  -0.04795 (0.01257) \\  
		proposed (gaussian, with LOFT) &  -0.01207 (0.00144) &  -0.01696 (0.00277) &  -0.03073 (0.00651) \\  
		proposed (student, with LOFT) &  -0.01127 (0.00197) &  -0.01844 (0.00362) &  -0.0306 (0.01334) \\  
		proposed (gaussian, no LOFT) &  -0.01031 (0.00111) &  -0.01754 (0.00449) &  \textbf{-0.02604} (0.01032) \\  
		proposed (student, no LOFT) &  -0.00979 (0.00207) &  -0.01734 (0.00476) &  -0.03379 (0.01077) \\  
		SMC &  -0.01938 (0.01633) &  -0.14798 (0.09068) &  -0.98769 (0.39748) \\
		\midrule 
		\multicolumn{4}{c}{  Multivariate Gaussian Mixture  } \\ 
		\midrule 
		\bfseries Method & \bfseries $d = 10$ & \bfseries $d = 100$ & \bfseries $d = 1000$ \\  
		\midrule 
		true value & 0.0 & 0.0 & 0.0  \\ 
		\midrule
		mean field gaussian &  -0.93939 (0.23324) &  -0.89837 (0.29501) &  -0.90655 (0.26212) \\  
		standard &  0.00003 (0.0012) &  -0.00063 (0.00182) &  -0.00015 (0.00529) \\  
		ATAF &  -0.00016 (0.00092) &  0.00111 (0.00343) &  0.00012 (0.007) \\  
		SymClip &  0.00004 (0.00116) &  -0.00179 (0.00194) &  -0.00073 (0.00512) \\  
		proposed (gaussian, with LOFT) &  \textbf{-0.0} (0.00115) &  -0.00083 (0.00262) &  \textbf{0.00007} (0.00818) \\  
		proposed (student, with LOFT) &  -0.00032 (0.00093) &  -0.00196 (0.00337) &  0.00229 (0.01066) \\  
		proposed (gaussian, no LOFT) &  -0.0 (0.00115) &  -0.00083 (0.00262) &  0.00007 (0.00818) \\  
		proposed (student, no LOFT) &  -0.00012 (0.00113) &  -0.00022 (0.00293) &  -0.00135 (0.00625) \\  
		SMC &  -0.00035 (0.00177) &  \textbf{-0.00019} (0.00257) &  -0.00046 (0.0023) \\
		\midrule 
		\multicolumn{4}{c}{  Conjugate Linear Regression  } \\ 
		\midrule 
		\bfseries Method & \bfseries $d = 11$ & \bfseries $d = 101$ & \bfseries $d = 1001$ \\  
		\midrule 
		true value & -269.73904 & -347.81384 & -320.59267  \\ 
		\midrule
		mean field gaussian &  -271.18867 (0.48871) &  -377.48859 (1.99237) &  -2096.65892 (15.65285) \\  
		standard &  -269.73934 (0.00054) &  \textbf{-347.81378} (0.00127) &  -321.12109 (0.27435) \\  
		ATAF &  -269.73912 (0.00026) &  -347.81345 (0.00242) &  -321.00036 (0.58395) \\  
		SymClip &  -269.73882 (0.00038) &  -347.81441 (0.00228) &  \textbf{-320.86489} (0.57739) \\  
		proposed (gaussian, with LOFT) &  -269.73879 (0.00038) &  -347.81429 (0.00213) &  -321.45662 (0.43597) \\  
		proposed (student, with LOFT) &  \textbf{-269.73907} (0.00041) &  -347.81414 (0.00249) &  -320.99025 (1.28594) \\  
		proposed (gaussian, no LOFT) &  -269.73879 (0.00038) &  -347.81429 (0.00213) &  -321.45662 (0.43597) \\  
		proposed (student, no LOFT) &  -269.73907 (0.00041) &  -347.81414 (0.00249) &  -320.99025 (1.28594) \\  
		SMC &  -281.18175 (3.00131) &  -595.63322 (22.78467) &  -3480.17078 (123.3412) \\
		\bottomrule 
	\end{tabular}
\end{table*}

\begin{table*}
	\centering
	\caption{Evaluation of all methods in terms of ELBO (standard deviation in brackets) for $d \in \{10, 100, 1000\}$; 
		number of flows $r = 16$.} \label{tab:ELBO_comparison_basic_models_r16}
	\footnotesize
	\begin{tabular}{llll}
		\toprule 
		\multicolumn{4}{c}{  Funnel  } \\ 
		\midrule 
		\bfseries Method & \bfseries $d = 10$ & \bfseries $d = 100$ & \bfseries $d = 1000$ \\  
		\midrule 
		mean field gaussian &  -1.86318 (0.00707) &  -3.0504 (0.00729) &  -4.20619 (0.00786) \\  
		standard &  -0.02689 (0.00183) &  \textbf{-0.04959} (0.00198) &  -0.25683 (0.0073) \\  
		ATAF &  \textbf{-0.01924} (0.00171) &  -0.05021 (0.0017) &  \textbf{-0.2305} (0.00641) \\  
		SymClip &  -0.42434 (0.00289) &  -0.84732 (0.00232) &  -1.03967 (0.00336) \\  
		proposed (gaussian, with LOFT) &  -0.02236 (0.00114) &  -0.10715 (0.00125) &  -0.38719 (0.00264) \\  
		proposed (student, with LOFT) &  -0.02472 (0.00074) &  -0.14684 (0.0019) &  -0.41061 (0.00382) \\  
		proposed (gaussian, no LOFT) &  -0.02236 (0.00114) &  -0.10715 (0.00125) &  -0.38719 (0.00264) \\  
		proposed (student, no LOFT) &  -0.02472 (0.00074) &  -0.14684 (0.0019) &  -0.41061 (0.00382) \\
		\midrule 
		\multicolumn{4}{c}{  Multivariate Student-T  } \\ 
		\midrule 
		\bfseries Method & \bfseries $d = 10$ & \bfseries $d = 100$ & \bfseries $d = 1000$ \\  
		\midrule 
		mean field gaussian &  -2.16601 (0.00651) &  -4.5299 (0.00767) &  -7.35684 (0.00516) \\  
		standard &  -0.03773 (0.00216) &  \textbf{-0.16146} (0.00566) &  -1.08358 (0.00683) \\  
		ATAF &  \textbf{-0.03397} (0.00147) &  -0.19026 (0.00388) &  -1.09653 (0.00962) \\  
		SymClip &  -0.2069 (0.00163) &  -0.52876 (0.00305) &  -1.58265 (0.00651) \\  
		proposed (gaussian, with LOFT) &  -0.07184 (0.00096) &  -0.23981 (0.00259) &  \textbf{-0.812} (0.00458) \\  
		proposed (student, with LOFT) &  -0.06782 (0.00093) &  -0.2354 (0.00199) &  -0.89013 (0.0052) \\  
		proposed (gaussian, no LOFT) &  -0.07184 (0.00096) &  -0.23981 (0.00259) &  -0.812 (0.00458) \\  
		proposed (student, no LOFT) &  -0.06782 (0.00093) &  -0.2354 (0.00199) &  -0.89013 (0.0052) \\
		\midrule 
		\multicolumn{4}{c}{  Multivariate Gaussian Mixture  } \\ 
		\midrule 
		\bfseries Method & \bfseries $d = 10$ & \bfseries $d = 100$ & \bfseries $d = 1000$ \\  
		\midrule 
		mean field gaussian &  -1.09054 (0.00072) &  -1.09085 (0.00086) &  -1.09121 (0.00081) \\  
		standard &  -0.03246 (0.00167) &  \textbf{-0.05054} (0.00246) &  -0.22116 (0.00461) \\  
		ATAF &  -0.01976 (0.00173) &  -0.06112 (0.00237) &  -0.19273 (0.00392) \\  
		SymClip &  -0.02465 (0.00143) &  -0.07069 (0.00255) &  \textbf{-0.17468} (0.00528) \\  
		proposed (gaussian, with LOFT) &  -0.01937 (0.00119) &  -0.08139 (0.00244) &  -0.19073 (0.00416) \\  
		proposed (student, with LOFT) &  \textbf{-0.01625} (0.00132) &  -0.06934 (0.00283) &  -0.21368 (0.00352) \\  
		proposed (gaussian, no LOFT) &  -0.01937 (0.00119) &  -0.08139 (0.00244) &  -0.19073 (0.00416) \\  
		proposed (student, no LOFT) &  -0.01625 (0.00132) &  -0.06934 (0.00283) &  -0.21368 (0.00352) \\
		\midrule 
		\multicolumn{4}{c}{  Conjugate Linear Regression  } \\ 
		\midrule 
		\bfseries Method & \bfseries $d = 11$ & \bfseries $d = 101$ & \bfseries $d = 1001$ \\  
		\midrule 
		mean field gaussian &  -275.82849 (0.0274) &  -399.32425 (0.05422) &  -2348.47948 (0.61318) \\  
		standard &  -269.7403 (0.00028) &  -347.84904 (0.00169) &  -332.96779 (0.04609) \\  
		ATAF &  \textbf{-269.74006} (0.00035) &  -347.86773 (0.00228) &  -331.99212 (0.03487) \\  
		SymClip &  -269.74247 (0.00049) &  -347.9524 (0.00331) &  -416.32816 (0.09648) \\  
		proposed (gaussian, with LOFT) &  -269.7403 (0.00037) &  -347.85343 (0.00186) &  \textbf{-331.47248} (0.02527) \\  
		proposed (student, with LOFT) &  -269.74073 (0.00047) &  \textbf{-347.84652} (0.00161) &  -331.62923 (0.03325) \\  
		proposed (gaussian, no LOFT) &  -269.7403 (0.00037) &  -347.85343 (0.00186) &  -331.47248 (0.02527) \\  
		proposed (student, no LOFT) &  -269.74073 (0.00047) &  -347.84652 (0.00161) &  -331.62923 (0.03325) \\
		\bottomrule 
	\end{tabular}
\end{table*}

\begin{table*}
	\centering
	\caption{Estimated log marginal likelihood (standard deviation in brackets) for $d \in \{10, 100, 1000\}$; 
		number of flows $r = 16$.} \label{tab:IS_comparison_basic_models_r16}
	\footnotesize
	\begin{tabular}{llll}
		\toprule 
		\multicolumn{4}{c}{  Funnel  } \\ 
		\midrule 
		\bfseries Method & \bfseries $d = 10$ & \bfseries $d = 100$ & \bfseries $d = 1000$ \\  
		\midrule 
		true value & 0.0 & 0.0 & 0.0  \\ 
		\midrule
		mean field gaussian &  -0.9051 (0.25602) &  -1.93821 (0.67967) &  -3.11768 (0.52003) \\  
		standard &  \textbf{0.00018} (0.01105) &  -0.00295 (0.00439) &  -0.03178 (0.02555) \\  
		ATAF &  -0.00471 (0.00219) &  \textbf{-0.00002} (0.00672) &  \textbf{-0.01926} (0.01717) \\  
		SymClip &  -0.23247 (0.03862) &  -0.66797 (0.05036) &  -0.86603 (0.06121) \\  
		proposed (gaussian, with LOFT) &  -0.01075 (0.00345) &  -0.0732 (0.00722) &  -0.2795 (0.01588) \\  
		proposed (student, with LOFT) &  -0.00391 (0.02007) &  -0.07329 (0.08636) &  -0.29113 (0.02494) \\  
		proposed (gaussian, no LOFT) &  -0.01075 (0.00345) &  -0.0732 (0.00722) &  -0.2795 (0.01588) \\  
		proposed (student, no LOFT) &  -0.00391 (0.02007) &  -0.07329 (0.08636) &  -0.29113 (0.02494) \\  
		SMC &  -0.1429 (0.06911) &  -1.76495 (0.36939) &  -3.50737 (0.46121) \\
		\midrule 
		\multicolumn{4}{c}{  Multivariate Student-T  } \\ 
		\midrule 
		\bfseries Method & \bfseries $d = 10$ & \bfseries $d = 100$ & \bfseries $d = 1000$ \\  
		\midrule 
		true value & 0.0 & 0.0 & 0.0  \\ 
		\midrule
		mean field gaussian &  -0.92496 (0.16171) &  -2.87842 (0.3654) &  -5.89942 (0.22141) \\  
		standard &  \textbf{-0.01292} (0.00612) &  -0.01293 (0.05005) &  -0.21087 (0.06764) \\  
		ATAF &  -0.01524 (0.00282) &  \textbf{-0.00086} (0.15296) &  \textbf{-0.17079} (0.07035) \\  
		SymClip &  -0.11578 (0.02809) &  -0.27219 (0.1134) &  -0.80163 (0.14787) \\  
		proposed (gaussian, with LOFT) &  -0.04565 (0.01678) &  -0.13769 (0.021) &  -0.22575 (0.04374) \\  
		proposed (student, with LOFT) &  -0.03505 (0.02326) &  -0.13828 (0.02616) &  -0.25835 (0.08885) \\  
		proposed (gaussian, no LOFT) &  -0.04565 (0.01678) &  -0.13769 (0.021) &  -0.22575 (0.04374) \\  
		proposed (student, no LOFT) &  -0.03505 (0.02326) &  -0.13828 (0.02616) &  -0.25835 (0.08885) \\  
		SMC &  -0.01938 (0.01633) &  -0.14798 (0.09068) &  -0.98769 (0.39748) \\
		\midrule 
		\multicolumn{4}{c}{  Multivariate Gaussian Mixture  } \\ 
		\midrule 
		\bfseries Method & \bfseries $d = 10$ & \bfseries $d = 100$ & \bfseries $d = 1000$ \\  
		\midrule 
		true value & 0.0 & 0.0 & 0.0  \\ 
		\midrule
		mean field gaussian &  -0.93939 (0.23324) &  -0.89837 (0.29501) &  -0.90655 (0.26212) \\  
		standard &  -0.00019 (0.00186) &  -0.00066 (0.00174) &  -0.00239 (0.01083) \\  
		ATAF &  0.00012 (0.0013) &  \textbf{0.00009} (0.00291) &  0.00243 (0.00582) \\  
		SymClip &  0.00058 (0.00226) &  -0.00059 (0.00267) &  -0.00351 (0.00885) \\  
		proposed (gaussian, with LOFT) &  \textbf{-0.0001} (0.00133) &  -0.00123 (0.00284) &  0.00088 (0.01197) \\  
		proposed (student, with LOFT) &  -0.00012 (0.0015) &  0.00109 (0.00353) &  -0.00217 (0.00491) \\  
		proposed (gaussian, no LOFT) &  -0.0001 (0.00133) &  -0.00123 (0.00284) &  0.00088 (0.01197) \\  
		proposed (student, no LOFT) &  -0.00012 (0.0015) &  0.00109 (0.00353) &  -0.00217 (0.00491) \\  
		SMC &  -0.00035 (0.00177) &  -0.00019 (0.00257) &  \textbf{-0.00046} (0.0023) \\
		\midrule 
		\multicolumn{4}{c}{  Conjugate Linear Regression  } \\ 
		\midrule 
		\bfseries Method & \bfseries $d = 11$ & \bfseries $d = 101$ & \bfseries $d = 1001$ \\  
		\midrule 
		true value & -269.73904 & -347.81384 & -320.59267  \\ 
		\midrule
		mean field gaussian &  -271.18867 (0.48871) &  -377.48859 (1.99237) &  -2096.65892 (15.65285) \\  
		standard &  \textbf{-269.73904} (0.00041) &  -347.81354 (0.00239) &  -321.87012 (0.83609) \\  
		ATAF &  -269.73903 (0.00027) &  \textbf{-347.81391} (0.00234) &  -321.86561 (0.4688) \\  
		SymClip &  -269.73905 (0.00056) &  -347.81312 (0.00791) &  -377.06485 (3.65637) \\  
		proposed (gaussian, with LOFT) &  -269.73924 (0.00038) &  -347.81352 (0.0013) &  \textbf{-321.66793} (0.96678) \\  
		proposed (student, with LOFT) &  -269.73885 (0.00037) &  -347.81321 (0.00196) &  -321.71383 (0.46365) \\  
		proposed (gaussian, no LOFT) &  -269.73924 (0.00038) &  -347.81352 (0.0013) &  -321.66793 (0.96678) \\  
		proposed (student, no LOFT) &  -269.73885 (0.00037) &  -347.81321 (0.00196) &  -321.71383 (0.46365) \\  
		SMC &  -281.18175 (3.00131) &  -595.63322 (22.78467) &  -3480.17078 (123.3412) \\
		\bottomrule 
	\end{tabular}
\end{table*}

\begin{table*}
	\centering
	\caption{Evaluation of all methods in terms of ELBO and log marginal likelihood estimates (standard deviation in brackets) for $d' \in \{10, 100, 1000\}$. } \label{tab:ELBO_and_IS_evaluation_horeshoe_synthetic_r64_r16}
	\footnotesize
	\begin{tabular}{llll}
		\toprule 
		\multicolumn{4}{c}{  Horseshoe Logistic Regression Model ($r = 64$) -- Synthetic Data -- ELBO  } \\ 
		\midrule 
		\bfseries Method & \bfseries $d' = 10$ & \bfseries $d' = 100$ & \bfseries $d' = 1000$ \\  
		\midrule 
		mean field gaussian &  -50.16087 (0.02565) &  -78.82118 (0.03113) &  -206.80469 (0.04076) \\  
		standard &  -43.58686 (0.00193) &  -56.82198 (1.10275) &  -170.82029 (0.39773) \\  
		ATAF &  -43.58192 (0.00146) &  -56.6599 (0.01174) &  -150.64962 (0.36023) \\  
		SymClip &  -43.6015 (0.00149) &  -56.63336 (0.01441) &  -154.68972 (0.06192) \\  
		proposed (gaussian, with LOFT) &  -43.57562 (0.00114) &  -56.16784 (0.00936) &  -145.28784 (0.06735) \\  
		proposed (student, with LOFT) &  -43.57637 (0.00101) &  -56.10268 (0.01114) &  \textbf{-105.29181} (0.07462) \\  
		proposed (gaussian, no LOFT) &  -43.59333 (0.00204) &  -56.1108 (0.00909) &  -156.21152 (0.07304) \\  
		proposed (student, no LOFT) &  \textbf{-43.57314} (0.00116) &  \textbf{-56.0003} (0.01031) &  -123.24537 (0.08508) \\
		\midrule 
		\multicolumn{4}{c}{  Horseshoe Logistic Regression Model ($r = 64$) -- Synthetic Data -- Log Marginal Likelihood Estimate  } \\ 
		\midrule 
		\bfseries Method & \bfseries $d' = 10$ & \bfseries $d' = 100$ & \bfseries $d' = 1000$ \\  
		\midrule 
		mean field gaussian &  -45.81527 (0.54777) &  -68.21041 (1.24219) &  -178.97184 (2.58527) \\  
		standard &  -43.55468 (0.00181) &  -54.70689 (0.33386) &  -143.3784 (1.98397) \\  
		ATAF &  \textbf{-43.55316} (0.00466) &  -54.73648 (0.58386) &  -122.16614 (2.35816) \\  
		SymClip &  -43.5537 (0.01553) &  -54.90284 (0.13389) &  -127.47038 (2.08928) \\  
		proposed (gaussian, with LOFT) &  -43.55563 (0.00156) &  -54.73608 (0.21571) &  -116.7371 (2.77658) \\  
		proposed (student, with LOFT) &  -43.55656 (0.00151) &  -54.70299 (0.16654) &  \textbf{-81.45998} (2.62084) \\  
		proposed (gaussian, no LOFT) &  -43.55392 (0.00389) &  -54.72848 (0.23261) &  -127.96218 (2.46497) \\  
		proposed (student, no LOFT) &  -43.5535 (0.00213) &  \textbf{-54.69696} (0.12413) &  -97.01509 (2.01398) \\  
		SMC &  -72.15519 (0.60622) &  -316.74614 (1.23563) &  -2605.75083 (0.82943) \\
		\midrule 
		\midrule
		\multicolumn{4}{c}{  Horseshoe Logistic Regression Model ($r = 16$) -- Synthetic Data -- ELBO  } \\ 
		\midrule 
		\bfseries Method & \bfseries $d' = 10$ & \bfseries $d' = 100$ & \bfseries $d' = 1000$ \\  
		\midrule 
		mean field gaussian &  -50.16087 (0.02565) &  -78.82118 (0.03113) &  -206.80469 (0.04076) \\  
		standard &  \textbf{-43.59642} (0.00235) &  -58.35805 (0.02742) &  -172.75058 (0.0779) \\  
		ATAF &  -43.60519 (0.00209) &  -58.47138 (0.10144) &  -151.93924 (0.17945) \\  
		SymClip &  -43.71616 (0.00359) &  -59.93733 (0.01882) &  -171.2301 (0.07379) \\  
		proposed (gaussian, with LOFT) &  -43.642 (0.00259) &  -58.80727 (0.01738) &  -166.96998 (0.06232) \\  
		proposed (student, with LOFT) &  -43.61169 (0.00202) &  -57.26398 (0.01502) &  \textbf{-129.22909} (0.08087) \\  
		proposed (gaussian, no LOFT) &  -43.64326 (0.00271) &  -58.78684 (0.01471) &  -167.61975 (0.05924) \\  
		proposed (student, no LOFT) &  -43.60367 (0.00203) &  \textbf{-57.25689} (0.01509) &  -130.20364 (0.07451) \\
		\midrule 
		\multicolumn{4}{c}{  Horseshoe Logistic Regression Model ($r = 16$) -- Synthetic Data -- Log Marginal Likelihood Estimate  } \\ 
		\midrule 
		\bfseries Method & \bfseries $d' = 10$ & \bfseries $d' = 100$ & \bfseries $d' = 1000$ \\  
		\midrule 
		mean field gaussian &  -45.81527 (0.54777) &  -68.21041 (1.24219) &  -178.97184 (2.58527) \\  
		standard &  -43.55501 (0.0045) &  -55.21554 (0.26658) &  -145.56326 (2.11553) \\  
		ATAF &  -43.55648 (0.00472) &  -55.3366 (0.18181) &  -122.46094 (2.26795) \\  
		SymClip &  -43.57685 (0.01127) &  -56.1934 (0.27895) &  -144.30236 (2.41617) \\  
		proposed (gaussian, with LOFT) &  -43.56231 (0.01606) &  -55.68322 (0.24981) &  -140.06385 (1.82293) \\  
		proposed (student, with LOFT) &  -43.55693 (0.00564) &  -54.97434 (0.21206) &  \textbf{-101.48784} (2.69617) \\  
		proposed (gaussian, no LOFT) &  -43.56263 (0.01568) &  -55.6475 (0.36711) &  -140.41891 (2.16415) \\  
		proposed (student, no LOFT) &  \textbf{-43.55214} (0.02036) &  \textbf{-54.93189} (0.27887) &  -101.52184 (2.82329) \\  
		SMC &  -72.15519 (0.60622) &  -316.74614 (1.23563) &  -2605.75083 (0.82943) \\
		\bottomrule 
	\end{tabular}
\end{table*}

\clearpage

\bibliographystyle{plainnat}
\bibliography{../../../all_papers_bibliography_extended}

\end{document}